\documentclass[10pt,twocolumn,letterpaper]{article}

\usepackage{cvpr}              

\usepackage[dvipsnames]{xcolor}

\definecolor{cvprblue}{rgb}{0.21,0.49,0.74}
\usepackage[pagebackref,breaklinks,colorlinks,citecolor=cvprblue]{hyperref}
\usepackage{amsmath}
\usepackage{rotating}
\usepackage{subcaption,graphicx}
\usepackage{multirow}
\usepackage{diagbox} 
\usepackage{adjustbox}
\usepackage[accsupp]{axessibility} 

\newcommand{\rulesep}{\unskip\ \vrule\ }

\def\paperID{10208} 
\def\confName{CVPR}
\def\confYear{2024}

\title{WALT3D: Generating Realistic Training Data from Time-Lapse Imagery \\
for Reconstructing Dynamic Objects under Occlusion}

\author{
Khiem Vuong$^{1,*}$ \qquad N Dinesh Reddy$^{2, *}$ \qquad Robert Tamburo$^{1}$ \qquad Srinivasa G. Narasimhan$^{1}$ \vspace{2mm} \\
$^{1}$Carnegie Mellon University \qquad $^{2}$Amazon\\
\url{https://www.cs.cmu.edu/~walt3d}
}

\begin{document}

\twocolumn[{%
\renewcommand\twocolumn[1][]{#1}%
\maketitle
\vspace{-1cm}
\begin{center}
    \centering
    \captionsetup{type=figure}
    \includegraphics[width=\textwidth]{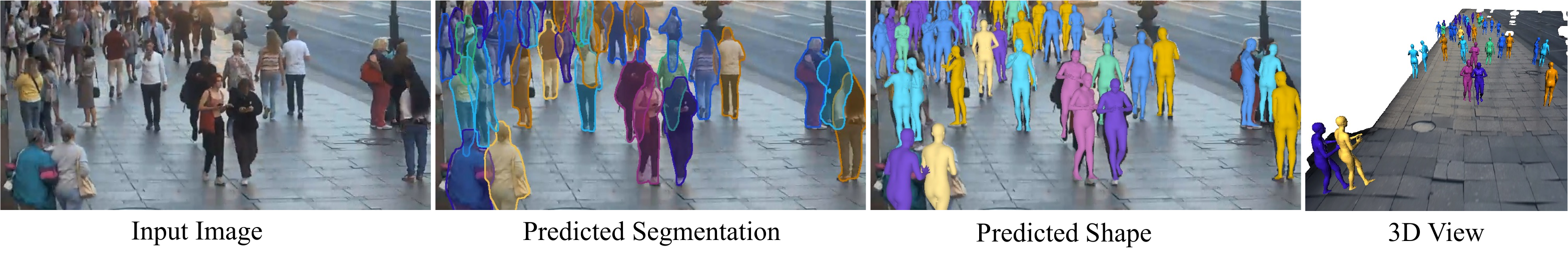}
    \vspace{-0.9cm}
    \captionof{figure}{Models trained on our automatically generated data from time-lapse imagery can reliably estimate amodal 2D bounding box, segmentation as well as 3D shape and pose despite the complex occlusions presented in the input image.}
\end{center}%
}]

{\let\thefootnote\relax\footnote{{$^*$denotes equal contribution and joint first author}}}


\begin{abstract}
Current methods for 2D and 3D object understanding struggle with severe occlusions in busy urban environments, partly due to the lack of large-scale labeled ground-truth annotations for learning occlusion.
In this work, we introduce a novel framework for automatically generating a large, realistic dataset of dynamic objects under occlusions using freely available time-lapse imagery. 
By leveraging off-the-shelf 2D (bounding box, segmentation, keypoint) and 3D (pose, shape) predictions as pseudo-groundtruth, unoccluded 3D objects are identified automatically and composited into the background in a clip-art style, ensuring realistic appearances and physically accurate occlusion configurations. 
The resulting clip-art image with pseudo-groundtruth enables efficient training of object reconstruction methods that are robust to occlusions.
Our method demonstrates significant improvements in both 2D and 3D reconstruction, particularly in scenarios with heavily occluded objects like vehicles and people in urban scenes.
\end{abstract}    
\section{Introduction}
\label{sec:intro}

In recent years, remarkable progress has been made in advancing scene understanding tasks such as object detection~\cite{ren2015faster, SWIN, li2022exploring}, tracking~\cite{xiang2015learning, zhang2022bytetrack, dendorfer2020mot20}, segmentation~\cite{MaskRCNN, cheng2021mask2former}, and 3D reconstruction~\cite{tulsiani2015viewpoints, cmrKanazawa18, kulkarni2020articulation}. These achievements are mainly attributed to the availability of large-scale datasets~\cite{lin2014microsoft, kitti, Xiang_Pascal, imagenet_cvpr09, deitke2023objaverse} and architectural innovations~\cite{krizhevsky2012imagenet, dosovitskiy2020image, he2016deep}. Despite this progress, a notable challenge persists in scenarios with severe \textit{occlusion}, where only a portion of the object is visible: an object may be partially occluded by other objects or truncated by the camera's field-of-view. This phenomenon is referred to as \textit{amodal perception}~\cite{kanizsa1979organization}, and it is hard to intuitively infer their complete shapes. Overcoming these challenges is important to advance many smart cities applications as well as robotics applications, where the number of cameras on vehicles and city infrastructure is rapidly increasing~\cite{glasl2008video, Naphade22AIC22, aslam2022detecting}.

Efforts to learn holistic representations necessitate a substantially annotated and realistic dataset. While recent works such as KINS~\cite{qi2019amodal}, COCO-Amodal~\cite{zhu2017semantic}, and Ithaca365~\cite{diaz2022ithaca365} have contributed by annotating some amodal ground-truth data, the available annotations remain limited. 
%
%
This scarcity is primarily due to the inherent difficulty in obtaining supervision for amodal representations, as labeling hidden parts of objects is a difficult task for people to  accomplish consistently~\cite{qi2019amodal, Reddy_2019_CVPR, zhu2017semantic}.
Moreover, 3D annotations under occlusions, including object shape and pose~\cite{krull2015learning, calli2017yale, wang2019normalized}, are even meager due to the difficulty of annotating 3D data, posing  challenges for 3D prediction tasks.

Expanding on WALT~\cite{Reddy_2022_CVPR}, we utilize time-lapse videos from stationary cameras to synthesize realistic occlusion scenarios by extracting unoccluded objects and composite them back into the background image at their original positions. 
Unlike WALT which focuses solely on compositing and learning 2D tasks, our approach extends to generating high-quality 3D pseudo-groundtruth data for robust 3D object reconstruction under occlusion.
Additionally, our 3D-based compositing method, named \textbf{WALT3D}, in contrast to WALT's simplistic 2D compositing, produces physically accurate occlusion configurations, leading to increased training data efficiency and scalability.

We start with the observation that, although not perfect, existing off-the-shelf methods demonstrate good accuracy in both 2D (segmentation~\cite{MaskRCNN}, keypoints~\cite{apo, sun2019deep}) and 3D (pose, shape~\cite{ke2020gsnet, wang2021nemo, ma2022robust}) prediction tasks, especially on unoccluded objects.
Thus, they can be used as \textit{``pseudo-groundtruth''} to improve the robustness of existing approaches in occlusion scenarios.
Next, we randomly select these unoccluded, non-intersecting 3D objects and put them back into the background image at their original positions (i.e., clip-art style). Specifically, we arrange them based on their distance from the camera, ensuring physically accurate and realistic occlusion configurations.
Each such resulting clip-art image is accompanied by amodal bounding box, segmentation masks, and 3D poses and shapes -- referred to as \textit{``pseudo-groundtruth''} as these were predicted by off-the-shelf methods on the original unoccluded objects.

Through extensive experiments, we demonstrate the effectiveness of our data in both vehicle and human reconstruction, particularly in scenarios with heavy occlusions.
It is important to note that our method does not require any human labeling and hence is easily scalable and serves as an effective method to automatically generate realistic training data for reconstructing dynamic objects under occlusion.

\noindent Our notable technical contributions are summarized below:
\begin{itemize}
  \item We introduce a novel method that automatically generates 2D/3D supervision data from time-lapse imagery with realistic occlusion configurations without human labeling.
  \item We demonstrate that the utilization of our generated data significantly enhances training efficiency and the accuracy of 2D/3D object reconstruction on real-world data, particularly in scenarios with high occlusion.
\end{itemize}

\section{Related Works}
\label{sec:related_works}
\noindent \textbf{Occlusion Reasoning:}
 Understanding and reasoning occlusions has been extensively studied for decades~\cite{fransens2006mean,gao2011segmentation,Schulter_2018_ECCV}. Bad predictions due to occlusions are dealt with as noise/outliers in robust estimators.  On the other hand, occlusions are explicitly treated as missing parts in model fitting methods~\cite{zhou20153d,vedaldi2009structured}. But severe occlusions, such as when a large part of an object is blocked, can result in poor model fitting~\cite{zia2015towards, gumeli2022roca}, especially when they attempt to simultaneously estimate the model fit as well as the missing parts. 
 
 

\noindent \textbf{2D Amodal Representation:}
Although the effects of occlusion on visual reasoning has been widely studied, estimating the amodal representation (i.e. both the occluded and visible regions) has only been recently explored. Initial attempts~\cite{zhu2017semantic, qi2019amodal, follmann2019learning, guo2012beyond} use a supervised learning paradigm using small datasets~\cite{zhu2017semantic, qi2019amodal} where humans have annotated occlusions to the best of their abilities. Some methods~\cite{Reddy_2018_CVPR, Reddy_2019_CVPR, simon2017hand} have explored using multiple views to provide accurate supervision for occluded parts but are not scalable due to capture limitations. To expand supervision, several methods synthesize occlusions to varying degrees of realism. But pure CG renderings~\cite{ehsani2018segan, alhaija2017augmented, li2017deep, fabbri2018learning, HuCVPR2019, Ke_2021_CVPR, Yuan_2021_CVPR} suffer from a wide domain-gap~\cite{krahenbuhl2018free, Schulter_2018_ECCV}. 
While driving datasets~\cite{kitti, argoverse, nuscenes, yu2020bdd100k} provide 2D amodal bounding boxes, annotating additional representations like segmentation and keypoints under occlusion remains challenging. Some methods, like Ghiasi et al.~\cite{ghiasi2021simple}, address this by randomly copy-pasting objects onto diverse background images. However, as these methods solely rely on 2D information, they generate unrealistic occlusion configurations leading to poor performance and limited training efficiency.




\noindent \textbf{3D Reconstruction Under Occlusion:}
  Reconstruction under severe occlusion is still in the nascent stages of research. Most algorithms developed focus on self-occluded objects with shape completion from partial observations~\cite{Zhou_2018_CVPR,choy20163d,park2019deepsdf}. On the other hand, shape models fitting for objects only with the visible regions either from images~\cite{3DRCNN_CVPR18,riegler2017octnet, ishimtsev2020cad, Reddy_2021_CVPR, gumeli2022roca} or depth sensors~\cite{agnew2020amodal, qi2017pointnet++, 7487799, wu2019pointconv} have been explored. Due to the inherent challenges in annotating 3D information, ground-truth data for 3D object understanding is limited, often confined to specific object sets~\cite{krull2015learning, wang2019normalized}, indoor~\cite{brazil2023omni3d}, or driving scenarios~\cite{kitti, nuscenes, apo}, and struggles to  generalize well to novel viewpoints such as stationary traffic cameras. Recent methods addressing pose estimation for vehicles~\cite{ma2022robust, wang2021nemo} and people~\cite{goel2023humans, kocabas2021pare} have demonstrated robustness in handling occlusion, either implicitly or explicitly. Thus, our dataset generation method can significantly enhance the robustness of these approaches to occlusion.

\section{Generating Realistic Supervision Data }
\label{sec:method_data_generation}
\begin{figure*}[t]
    \centering

\includegraphics[width=0.9\textwidth]{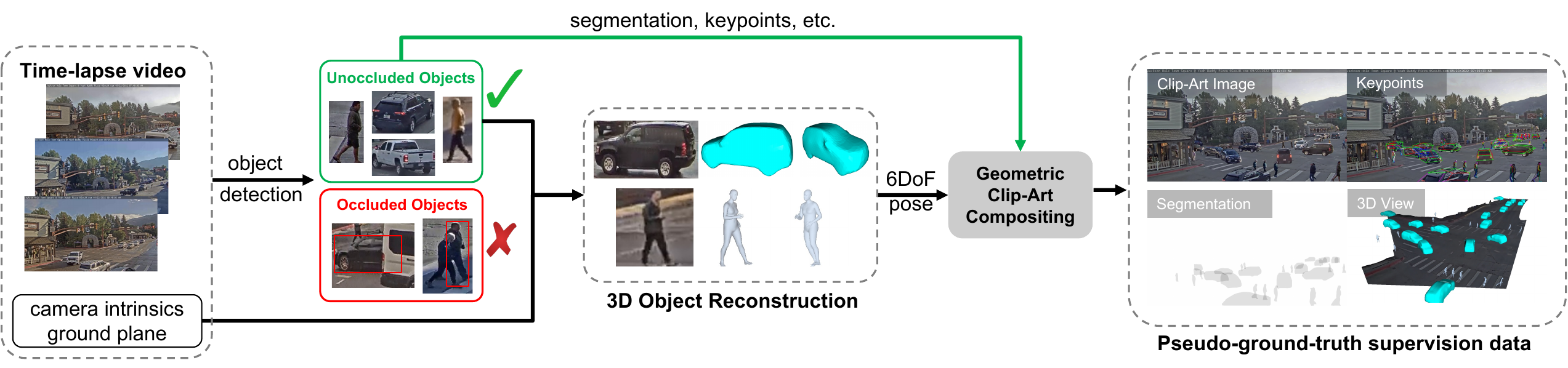}
\vspace{-0.1in}
\caption{Given a time-lapse video, we automatically generate 2D/3D training data under severe occlusions. We start by detecting each object in the video, and unoccluded (fully visible) objects are identified. Each unoccluded object is then reconstructed using the ground plane and camera parameters. With the 3D pose, unoccluded objects are composited back into the same location (i.e., clip-art style) in a geometrically consistent approach. The composited image and its pseudo-groundtruth from off-the-shelf methods (e.g., segmentation, keypoints, shapes) are utilized to train a model that can produce accurate 2D/3D object reconstruction under severe occlusions. }
    \label{fig:pipeline}
    \vspace{-0.15in}
\end{figure*}

An overview of our approach is shown in Figure~\ref{fig:pipeline}. From a time-lapse video, we  identify unoccluded objects and extract their 2D attributes such as segmentation and keypoints. We then use off-the-shelf 3D object reconstruction methods to obtain the pose and shape of these objects, constrained by the camera intrinsics and ground plane. After that, we re-insert these non-intersecting 3D objects into the background image at their original positions in a ``clip-art'' style, arranged based on their distance from the camera to ensure the occlusion configurations are physically accurate and realistic. Finally, we use the clip-art composited image together with its pseudo-groundtruth supervision data to learn robust 2D/3D object reconstruction under occlusion.

\noindent \textbf{Mining Unoccluded Objects:} Given a stream of time-lapse data from a camera, our goal is to mine for unoccluded objects. In this context, ``unoccluded'' denotes instances where a detected object is not obstructed by any other object or truncated out of the field-of-view. 
On the time-lapse feed from a camera, we run instance segmentation~\cite{SWIN} on each frame and use a simple object tracker~\cite{Wojke2017simple} to track the
detected bounding box and segmentation that belongs to ``person'' and ``car'' classes.  
%
The simplest heuristic for detecting unoccluded objects involves calculating the Intersection over Union (IOU) of detected objects. In previous methods like~\cite{Reddy_2022_CVPR}, objects were considered unoccluded if the intersection of their bounding box bottom with any other box was smaller than a threshold $\delta$. However, this strategy becomes less reliable in scenarios where dynamic objects are occluded by static elements, such as vehicles being occluded by buildings, trees, poles, etc.

%
%

%
To address this limitation, we train a simple Occlusion Classifier (OC) to categorize each detected object as either unoccluded or occluded, which is trained using a supervised approach with human annotators who label objects as unoccluded/occluded.
Despite the OC module outperforming the heuristic filter in classifying objects' occlusion status, we did not observe a significant improvement in downstream evaluations (less than $\pm0.5$ AP) due to the small amount of outliers overall in the training data.
Therefore, it is not a mandatory component in our pipeline: the simple heuristics proposed in WALT~\cite{Reddy_2022_CVPR} can be employed for the unoccluded object detection task without additional training.
Nevertheless, we believe this data can be useful for occlusion reasoning tasks and we will publicly release it.

%

\noindent \textbf{Reconstructing Unoccluded Objects}:
Once unoccluded objects are identified, we describe the 3D reconstruction process of two primary classes: vehicles and humans.

Each mined unoccluded vehicle is reconstructed following Li et al.~\cite{li2021traffic4d}. We parameterize each vehicle's 3D keypoints $\mathbf{X}$ by a linear combination of the mean shape $\mathbf{\bar{Q}}$ and $K$ principal components $\mathbf{Q}_1,\dots,\mathbf{Q}_K$ computed from an
object CAD model dataset~\cite{li2017deep}: $\mathbf{X} = \mathbf{\bar{Q}} + \sum_{k=1}^K \alpha_k \mathbf{Q}_k$, where $\alpha_k$ is the shape coefficient that needs to be optimized. Starting from the mean shape $\mathbf{\bar{Q}}$, we first detect the 2D keypoints for each unoccluded vehicle and initialize the 6-DoF poses using EPnP~\cite{lepetit2009epnp}. Subsequently, for each track of detected vehicles, we optimize for the 6-DoF object poses while regularizing the shape parameter $\alpha_k$ to be constant for the same vehicle in different frames by minimizing reprojection errors. Additionally, we also constrain the objects to lie on the constant global ground plane, ensuring physically plausible reconstructions.

In the context of human reconstruction, we employ the Human Mesh Recovery (HMR) method HMR 2.0 from Goel et al.~\cite{goel2023humans}. Using HMR 2.0, we predict the SMPL pose and shape parameters for each detected unoccluded human~\cite{SMPL:2015}.
To determine camera translation, we find the intersection between the backprojected ray from the bottom of the 2D bounding box with the ground plane. Given the camera's intrinsic matrix $\mathbf{K}$, ground plane equation $\begin{bmatrix}\mathbf{n} & d\end{bmatrix}$, and $\mathbf{p}_b$ as the pixel coordinate of the bottom of the box, its depth is computed as $z_b = -\frac{d}{\mathbf{n}^T(\mathbf{K}^{-1} \mathbf{p}_b)}$.
For camera intrinsic parameters and ground plane equation, we employ a recent method by Vuong et al.~\cite{vuong2024trafficcalib} that utilizes Google Street View~\cite{GoogleStreetView} to automatically calibrate stationary traffic cameras and infer the ground plane.
The mined unoccluded object image with its corresponding segmentation mask, keypoint locations, and 3D poses are used in a clip-art based framework to generate 2D and 3D supervision data in severe occlusions as illustrated in Fig.~\ref{fig:pipeline}.

\begin{figure*}[t]
    \centering

\includegraphics[width=0.9\textwidth]{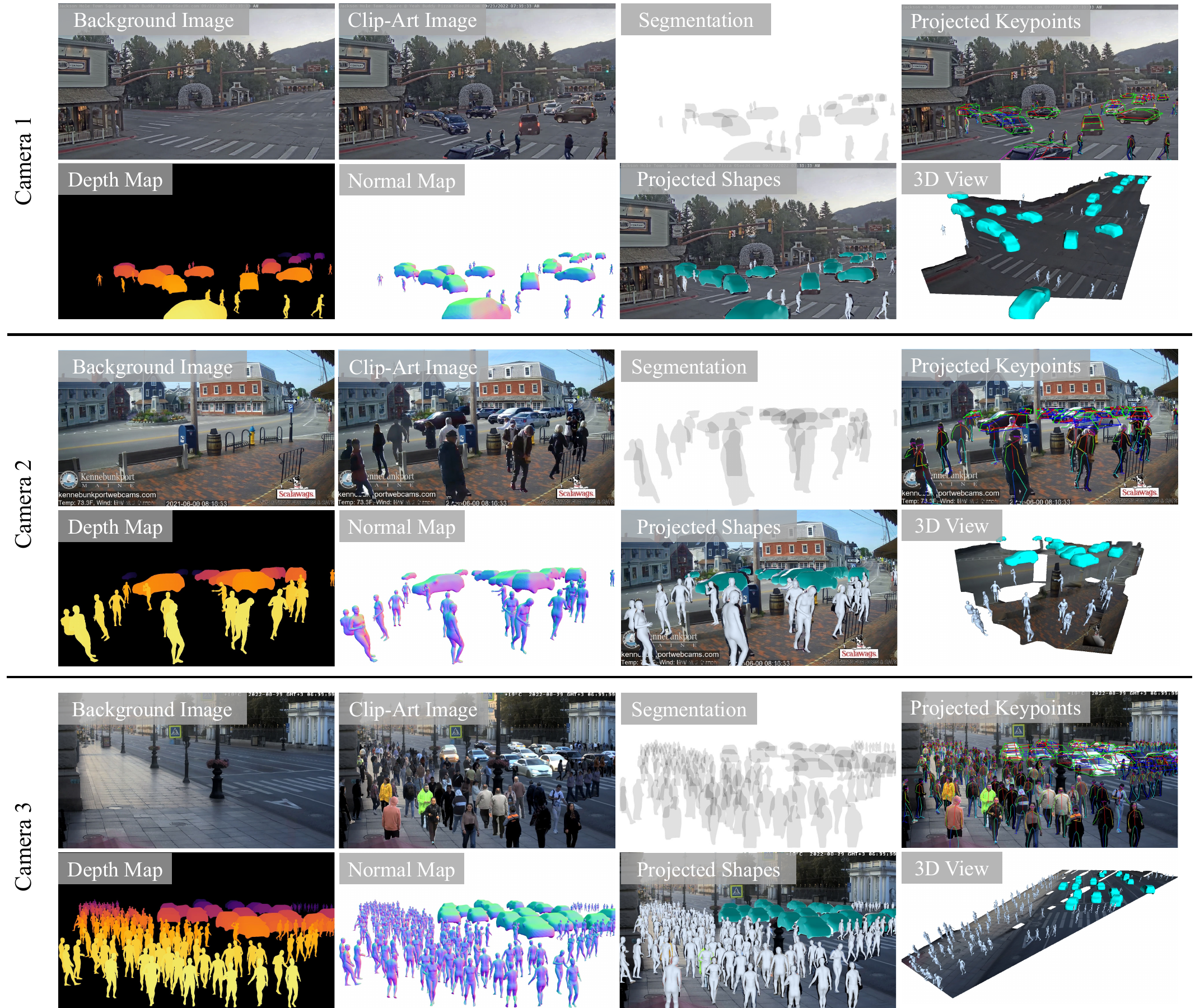}
\vspace{-0.08in}
\caption{{\bf Automatically generated 2D and 3D Clip-Art to supervise our network:} Unoccluded objects are first mined using time-lapse imagery of WALT dataset~\cite{Reddy_2022_CVPR}. Non-intersecting unoccluded objects are composited back into the background image in their respective original positions to preserve correct appearances. The resulting clip-art images, along with their corresponding amodal pseudo-groundtruth information, such as segmentation, keypoints, depth/normal maps, and 3D shapes, are shown. Our method generates realistic appearances from any stationary camera, incorporating diverse viewing geometries, weather conditions, lighting, and occlusion configurations.}
    \label{fig:clipart}
    \vspace{-0.15in}
\end{figure*}

\begin{figure*}[ht!]
    \centering
\includegraphics[width=0.85\textwidth]{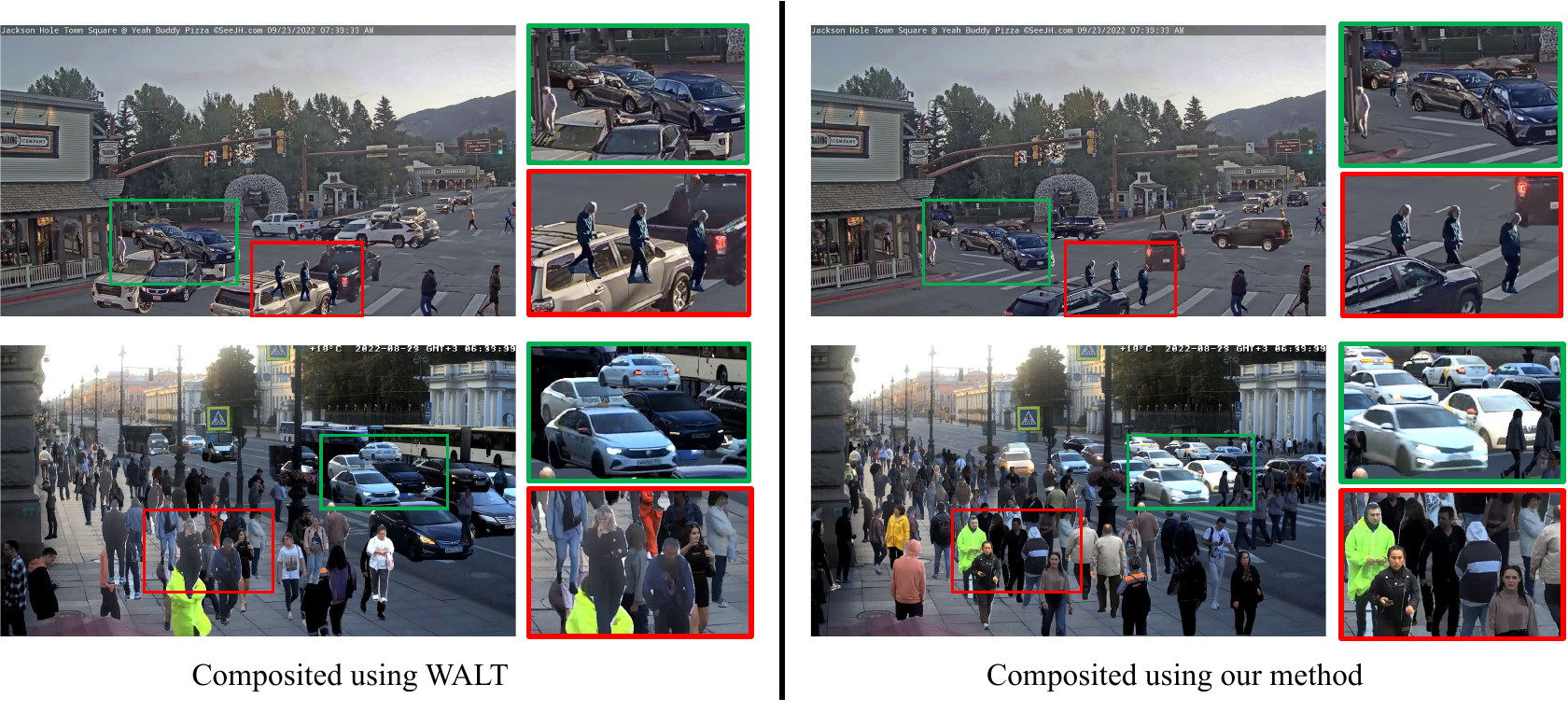}
\vspace{-0.15in}
\caption{Comparison between images composited using the 2D-based method WALT2D~\cite{Reddy_2022_CVPR} (left) and our 3D-based method WALT3D (right). It is evident that our 3D-based compositing method generates realistic and geometrically accurate occlusion configurations, in contrast to the 2D-based method (e.g., cars and people overlapping in an unfeasible way).}
    \label{fig:compare_walt2d}
    \vspace{-0.15in}
\end{figure*}

\noindent \textbf{Pseudo-groundtruth Data Generation:} 
To obtain supervision for occluded objects, we use image-based compositing to generate composited images, serving as input for training a network to learn 2D/3D reconstruction in severe occlusions.
During the compositing process, various 2D/3D representations for each unoccluded object, including keypoints, segmentation, pose, and shape obtained from off-the-shelf methods, serve as pseudo-groundtruth supervision signals even after compositing. 
By using the clip-art-based compositing approach described below, we automatically generate a large number of realistic supervision signals in severe occlusions.
Leveraging the 3D poses of mined unoccluded objects, we geometrically compose the object's image in a clip-art manner. Non-intersecting 3D objects are randomly sampled and reinserted into the background image (i.e., median image) at their original positions, ranging from the farthest to the closest. This approach ensures the creation of physically accurate and realistic occlusion configurations (see Fig.~\ref{fig:clipart}).
Crucially, our geometry-based approach distinguishes itself from the mere compositing of 2D images such as WALT~\cite{Reddy_2022_CVPR}, which can result in unrealistic composition (see the comparison in Fig.~\ref{fig:compare_walt2d}). Our data with physically accurate and realistic occlusion configurations contributes to more efficient and scalable training, as demonstrated in our experimental results.
Each such generated clip-art image is accompanied by amodal segmentation masks, keypoints, 3D poses and shapes, provided by off-the-shelf methods that serve as pseudo-groundtruth supervision to learn object reconstruction under occlusion.
%

\section{Learning Reconstruction under Occlusion}
\label{sec:network}
We have generated an extensive clip-art image dataset with corresponding 2D/3D pseudo-groundtruth representations (see Fig.~\ref{fig:clipart}). Using this supervision signal, we train a model capable of inferring holistic object representations in the presence of severe occlusions. 

\noindent \textbf{Base Network and 2D Amodal Representations:} Using the Swin Transformer~\cite{SWIN} backbone with MaskRCNN-based~\cite{MaskRCNN} detection heads for its simplicity, our goal is to show improvement irrespective of the base model used.
We train the default network with a multi-task loss $\mathcal{L}_\texttt{2D}$ for each representation. These losses, computed on each sampled Region of Interest (RoI), include label classification loss $\mathcal{L}_\texttt{cls}$, bounding-box loss $\mathcal{L}_\texttt{box}$, binary cross-entropy loss on the mask branch $\mathcal{L}_\texttt{mask}$ and keypoints $\mathcal{L}_\texttt{kp}$: $\mathcal{L}_\texttt{2D} = \mathcal{L}_\texttt{cls} + \mathcal{L}_\texttt{box} + \mathcal{L}_\texttt{mask} + \mathcal{L}_\texttt{kp}$. Crucially, with our clip-art dataset including \textit{amodal} 2D bounding box, segmentation, and keypoints, our method effectively learns \textit{amodal} instance segmentation and keypoint detection.

\noindent \textbf{3D Object Reconstruction:} To regress object 3D pose and shape under occlusion, we extend the base network with a 3D regression branch.
Each vehicle $N$ semantic 3D keypoints $\mathbf{X}$ are represented by a linear combination of the mean shape $\mathbf{\bar{Q}}$ and $K$ principal components $\mathbf{Q}_{1\dots K}$, where $\alpha_k$ is the shape coefficients: $\mathbf{X} = \mathbf{\bar{Q}} + \sum_{k=1}^K \alpha_k \mathbf{Q}_k$.
Thus, we want to regress for its 6-DoF pose $\{(\mathbf{R},\mathbf{t}) \in \mathbf{SE}(3)\}$ and shape coefficients $\alpha_k$.
Given the predicted 2D keypoints and the corresponding 3D keypoints $\mathbf{X}$, we can obtain its 6-DoF pose through a differentiable PnP layer~\cite{BPnP2020}. 
Defining $\pi(\cdot)$ as the projection function with known intrinsic $\mathbf{K}$, we can optimize for the keypoint reprojection loss $\mathcal{L}_{\texttt{kp2D}}$, shape loss $\mathcal{L}_{\texttt{shape}}$, and pose loss $\mathcal{L}_{\texttt{pose}}$ as:
\begin{equation}
\begin{aligned}
    \mathcal{L}_{\texttt{kp2D}} = \sum_{i=1}^N || \pi &(\mathbf{R} \mathbf{X}_i + \mathbf{t}) - \mathbf{x}_i^*||_2^2, \\
    \mathcal{L}_{\texttt{shape}} = || \alpha_k - \alpha_k^*||_2^2&,~  \mathcal{L}_{\texttt{pose}} = ||[\mathbf{R}, \mathbf{t}] - [\mathbf{R}^*, \mathbf{t}^*]||
\end{aligned}
\end{equation}
Here, $\mathbf{x}_i^*, \alpha_k^*, \mathbf{R}^*, \mathbf{t}^*$ represents the (pseudo)-groundtruth 2D keypoints, shape coefficients, rotation, and translation respectively. The final loss term is given as the weighted sum of the losses, and we learn amodal bounding box, segmentation, keypoint locations, 3D shape and pose in an end-to-end differentiable manner. 

For human reconstruction, we train our HMR predictor branch like HMR 2.0~\cite{goel2023humans}.
Supervised by pseudo-groundtruth SMPL pose $\theta^*$ and shape $\beta^*$ parameters~\cite{SMPL:2015}, our network is guided by the loss functions on SMPL parameters $\mathcal{L}_{\texttt{smpl}}$, 3D keypoints $\mathcal{L}_{\texttt{kp3D}}$, 2D keypoints $\mathcal{L}_{\texttt{kp2D}}$, and an adversarial loss $\mathcal{L}_{\texttt{adv}}$ that ensures the model predicts valid 3D poses through the discriminator $D_k$:
\begin{equation}
    \begin{aligned}
    \mathcal{L}_{\texttt{smpl}} = ||&[\theta, \beta] - [\theta^*, \beta^*]||_2^2,\\
    \mathcal{L}_{\texttt{kp3D}} = ||\mathbf{X} - \mathbf{X}^*||_1&,~ \mathcal{L}_{\texttt{kp2D}} = ||\pi(\mathbf{X}) - \mathbf{x}^*||_1, \\
    \mathcal{L}_{\texttt{adv}} = \sum_k & (D_k(\theta_b, \beta) - 1)^2 
\end{aligned}
\end{equation}

Here, $\mathbf{x}^*$ and $\mathbf{X}^*$ denote the (pseudo)-groundtruth 2D and 3D keypoints respectively. With a weighted combination of these losses, we optimize for human 3D shape and pose in an end-to-end fashion.

\section{Dataset and Implementation Details}
\label{sec:implementation}

\noindent \textbf{Raw Time-lapse Dataset~\cite{Reddy_2022_CVPR}:} This dataset contains images from 20 stationary cameras in urban scenes captured over multiple years. The images are either 4K or HD and are captured at 60fps in short bursts. We used 30 days of data from 10 cameras resulting in approximately 3.3 million car and people instances for our experiments.

\noindent \textbf{WALT2D~\cite{Reddy_2022_CVPR}:} From the raw time-lapse dataset, we use the 2D-based compositing method proposed by WALT~\cite{Reddy_2022_CVPR} to generate supervision data by pasting unoccluded objects back into the scene with different backgrounds resulting in 10000 training and 500 testing images per camera. Fig.~\ref{fig:compare_walt2d} (left) illustrates an example of this dataset. It is important to note that this compositing method, relying solely on 2D information, leads to intersecting objects in the real world that may not faithfully represent the physical configuration.

\noindent \textbf{WALT3D (Ours):} Also from the raw time-lapse dataset, our approach involves mining unoccluded objects and re-insering them into the scene, resulting in a comparable number of object instances as 2D-based compositing in WALT2D~\cite{Reddy_2022_CVPR}. However, our approach employs a 3D-based compositing method to generate supervision data, as outlined in Section~\ref{sec:method_data_generation}, in contrast to relying solely on 2D-based composition. Fig.~\ref{fig:compare_walt2d} (right) shows an example clip-art image composited using our method, with our approach yielding training data with greater realism and physical accuracy compared to WALT2D.


\begin{figure}
    \centering
        \includegraphics[width=0.9\columnwidth]{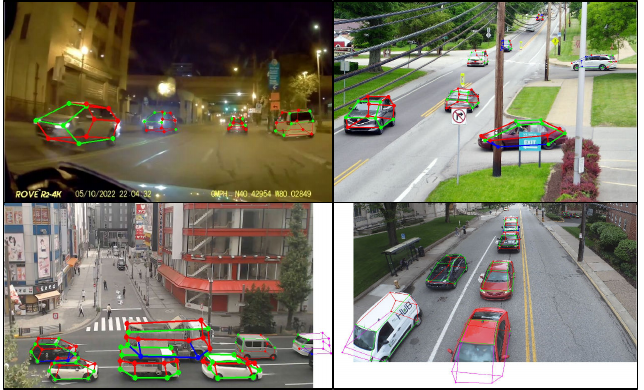}
    \vspace{-0.1in}
    \caption{Sample images from our new vehicle 2D keypoints dataset. The dataset contains a wide range of appearance variations including day and night and various traffic scenarios.}
    \label{fig:dataset}
\vspace{-0.2in}
\end{figure}


\noindent \textbf{Vehicle 2D Keypoints Dataset (Ours):} To determine 6-DoF vehicle pose during compositing, we require high-quality 2D vehicle keypoints. Existing datasets~\cite{Xiang_Pascal,li2017deep,Reddy_2018_CVPR,apo} offer human-annotated 2D vehicle keypoints but mainly focus on driving scenes or have limited training examples, lacking necessary appearance diversity for novel viewpoints like our stationary traffic cameras. Thus, we propose a new dataset with 7,018 images and 42,547 annotated instances from diverse viewpoints (see Fig.~\ref{fig:dataset}), each keypoint containing 2D location and occlusion status. With our dataset, we observe significant improvement in keypoint localization accuracy ($66.41\%$ to $80.12\%$ on PCK@0.1), and consequently, an improvement in pseudo-groundtruth data quality. Further details are in the Supplementary.

\section{Baselines, Metrics, and Evaluations}

\subsection{Baselines}
\noindent \textbf{Detection and Instance Segmentation:} All baselines share the same network architecture, employing a Swin~\cite{SWIN} backbone as detailed in Section~\ref{sec:network}. The \textbf{SWIN} baseline is initially trained on the COCO dataset~\cite{lin2014microsoft}. Subsequently, we perform further finetuning on the \textbf{WALT2D} and \textbf{WALT3D (Ours)} datasets, as detailed in Section~\ref{sec:implementation}.

\noindent \textbf{Vehicle 3D Reconstruction:} We compare our approach with \textbf{Occ-Net}~\cite{Reddy_2022_CVPR} and \textbf{3DRCNN}~\cite{3DRCNN_CVPR18} by changing their feature extraction backbone to be as close as possible to ours for fair comparison. They are trained on a combination of PASCAL3D+~\cite{Xiang_Pascal}, KITTI-3D~\cite{li2017deep}, Carfusion~\cite{Reddy_2018_CVPR}, and ApolloCar3D~\cite{apo}, where we further finetune our baselines on the \textbf{WALT2D} and our \textbf{WALT3D} dataset, respectively.

\noindent \textbf{Human 3D Reconstruction:} Our architecture mirrors HMR 2.0~\cite{goel2023humans} as described in Section~\ref{sec:network}. We initialize from the pretrained model, then finetune on our \textbf{WALT3D} data.

\subsection{Evaluation Metrics}
\noindent \textbf{2D Metrics:} We report Average Precision (AP)  on bounding box detection ($\text{AP}^{\text{box}}$) and instance segmentation ($\text{AP}^{\text{mask}}$) following~\cite{lin2014microsoft}. For keypoints, we use Percentage of Correct Keypoints (PCK) metric where a keypoint is considered correct if it lies within the radius $\alpha$ (with $0 < \alpha < 1$) of the ground-truth keypoint.

\noindent \textbf{3D Metrics:} We use the standard metrics of previous work~\cite{hmrKanazawa17}, reporting the Mean Per Joint Position Error (MPJME) between predicted and ground-truth 3D keypoints (aligned using the root joint) as well as 3D PCK.

\subsection{Ablation Analysis}

\begin{figure}[t!]
    \centering
        \includegraphics[width=0.23\textwidth]{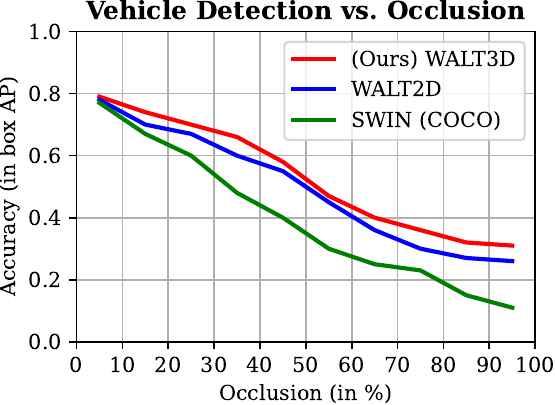}
        \vspace{0.3cm}
        \includegraphics[width=0.23\textwidth]{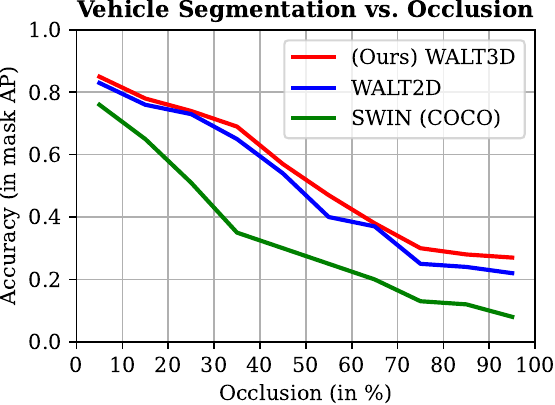}
        \includegraphics[width=0.23\textwidth]{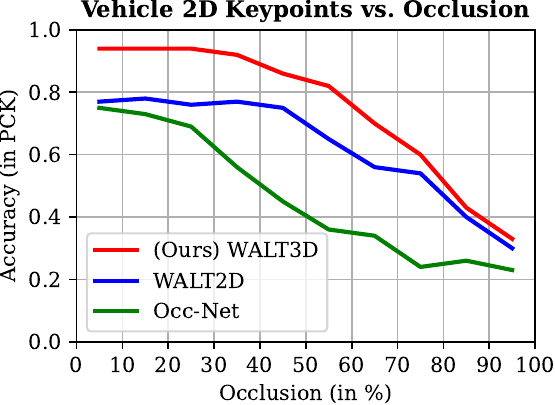}
        \includegraphics[width=0.23\textwidth]{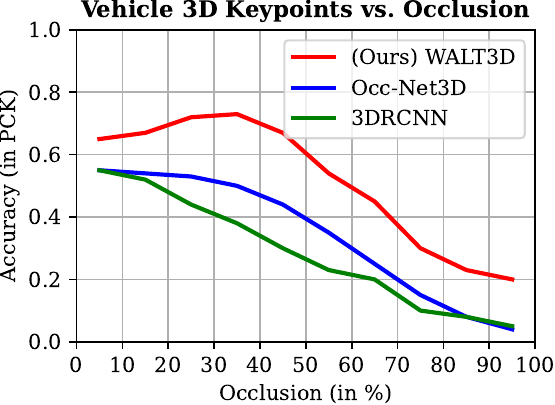}
    \vspace{-0.1in}
    \caption{We show the accuracy of our method with respect to increasing percentage of occlusion on multiple tasks like amodal vehicle detection, segmentation, 2D and 3D keypoint  estimation. Observe that our method consistently performs better than other baselines showing robustness to increasing occlusion percentage.}
    \label{fig:comparision}
    \vspace{-0.1in}

\end{figure}

\begin{figure}[t!]
    \centering
        \includegraphics[width=0.23\textwidth]{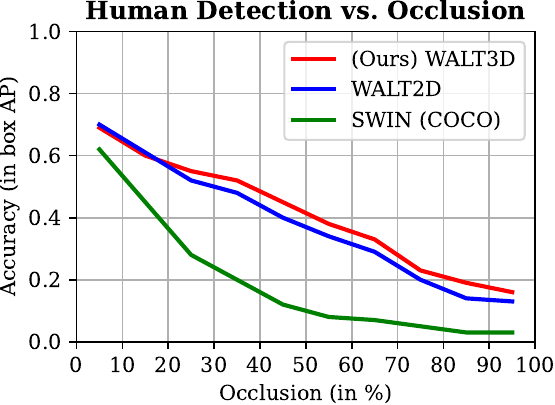}
        \vspace{0.2cm}
        \includegraphics[width=0.23\textwidth]{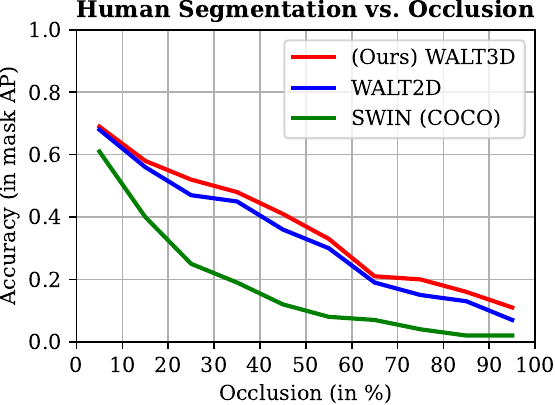}
        \includegraphics[width=0.23\textwidth]{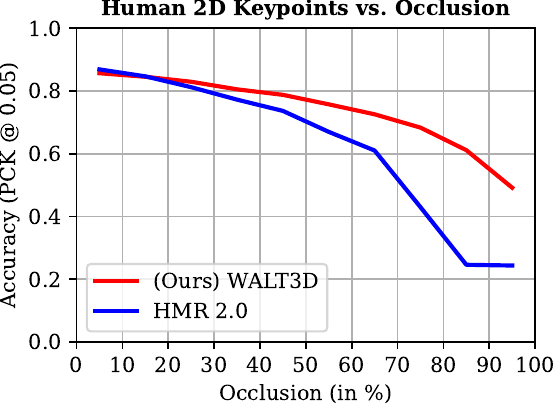}
        \includegraphics[width=0.23\textwidth]{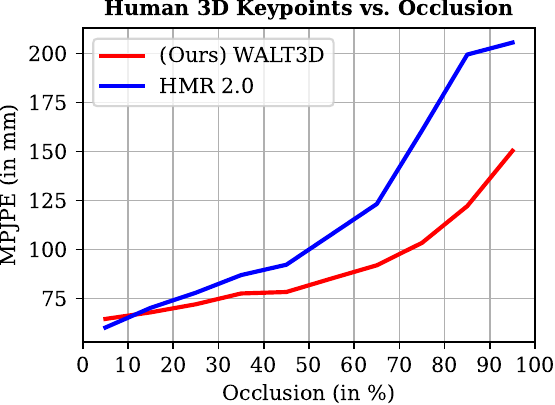}
    \vspace{-0.1in}
    \caption{We show the accuracy of our method with respect to increasing percentage of occlusion on multiple tasks like amodal human detection, segmentation, and human mesh recovery (HMR). Observe that our method consistently performs better than other baselines showing robustness to increasing occlusion percentage.}
    \label{fig:comparision_human}
    \vspace{-0.2in}

\end{figure}

\begin{figure}[t!]
    \centering
        \includegraphics[width=0.9\columnwidth]{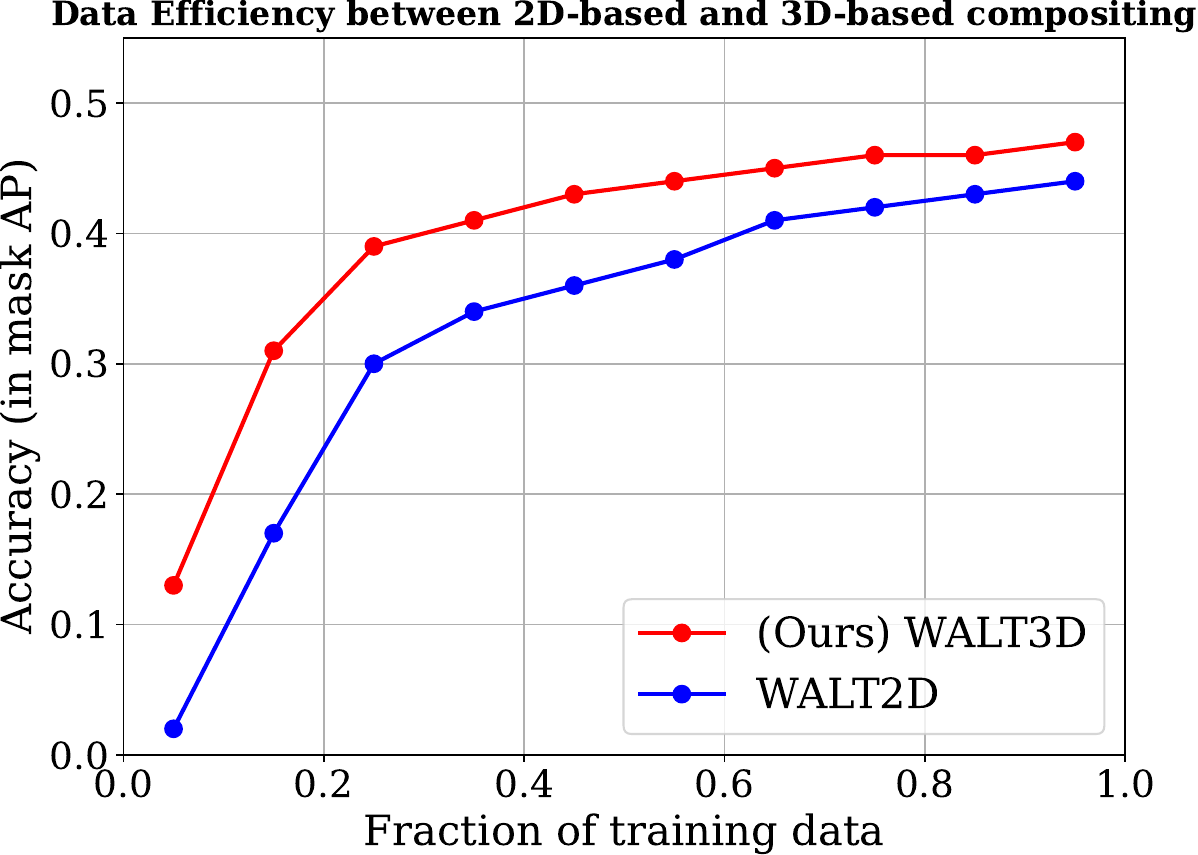}
        \vspace{-0.1in}
    \caption{Comparison between WALT2D~\cite{Reddy_2022_CVPR}) and our WALT3D approach. Since our method produces  higher quality training data with more realistic occlusion configuration, our approach is especially useful in low-data regime. }
    \label{fig:data_efficiency}
    \vspace{-0.2in}

\end{figure}

\noindent \textbf{Robustness to Occlusions:} We evaluate the effectiveness of our algorithm with varying occlusion levels. Similar to WALT~\cite{Reddy_2022_CVPR}, we use the pseudo-groundtruth segmentation masks from the evaluation set to group objects based on occlusion percentage. 
In 2D tasks such as detection and segmentation, the model trained with our \textbf{WALT3D} data significantly outperforms the baseline \textbf{SWIN} model trained on COCO~\cite{lin2014microsoft}, which only includes modal masks. This improvement is particularly evident in high occlusion percentages for both vehicle (Fig.~\ref{fig:comparision}) and human (Fig.~\ref{fig:comparision_human}). Compared to \textbf{WALT2D}~\cite{Reddy_2022_CVPR}, our data demonstrates slightly better accuracy across all occlusion levels, thanks to our 3D-based compositing method \textbf{WALT3D} producing physically accurate occlusion configurations with the same amount of training data.
For 3D tasks such as vehicle (Fig.~\ref{fig:comparision}) and human (Fig.~\ref{fig:comparision_human}) 3D pose estimation, using our \textbf{WALT3D} data consistently outperforms other baselines, particularly in high occlusion percentages.
Thus, models trained with our \textbf{WALT3D} data, even with imperfect pseudo-groundtruth from off-the-shelf methods, are robust to occlusion.

\noindent \textbf{3D-based Compositing helps Data Efficiency:} Fig.~\ref{fig:data_efficiency} shows that our \textbf{WALT3D} data significantly improves performance across all training data fractions, including at low data regime (at 15\% of training data with +13.9 $\text{AP}^{\text{mask}}$ improvement compared to training with \textbf{WALT2D} data). Essentially, a model trained on 25\% of \textbf{WALT3D} data achieves comparable AP to one trained on 60\% of \textbf{WALT2D} data. This improvement stems from our \textbf{WALT3D} approach generating higher quality and physically accurate occlusion configurations with the same training data volume, aligning more closely with the real-world distribution as in Fig.~\ref{fig:compare_walt2d}. Importantly, this data efficiency enables scaling to a larger number of scenes more efficiently.

\noindent{\bf Cross-evaluations on other datasets:} Table~\ref{tab:cocoa} presents quantitative results on COCO-Amodal~\cite{zhu2017semantic} (COCO-A) (natural images) and KINS~\cite{qi2019amodal} (driving images), each containing human-annotated ground-truth 2D amodal instance masks. It shows that additional fine-tuning with our \textbf{WALT3D} data improves results in both KINS and COCO-A, highlighting the cross-evaluation performance of our data, with potential for even better generalization with the increasing amount of automatically generated data.

\begin{table}[h]
\centering
\resizebox{\columnwidth}{!}{%
\begin{tabular}{l|cc|cc}
\toprule
\multirow{2}{*}{\diagbox{Training Data}{~~~~~~~~~~~Testing Data}} & \multicolumn{2}{c|}{Box AP}        & \multicolumn{2}{c}{Mask AP}        \\ \cline{2-5} 
                     & \multicolumn{1}{c}{KINS} & COCO-A & \multicolumn{1}{c}{KINS} & COCO-A \\ \midrule
COCO               & \multicolumn{1}{c}{27.8}     &  38.0   & \multicolumn{1}{c}{20.8}     &   31.2  \\
COCO + WALT2D        & \multicolumn{1}{c}{28.2}     &  40.4     & \multicolumn{1}{c}{22.4}     &    36.4 \\
\textbf{COCO + WALT3D}        & \multicolumn{1}{c}{\textbf{28.8}}     &  \textbf{43.2}     & \multicolumn{1}{c}{\textbf{23.5}}     &    \textbf{38.7} \\  \midrule
COCO + COCO-A + KINS              & \multicolumn{1}{c}{36.8}     &  46.9   & \multicolumn{1}{c}{32.9}     &   42.7  \\ 
COCO + COCO-A + KINS + WALT2D            & \multicolumn{1}{c}{37.4}     &  47.9   & \multicolumn{1}{c}{33.3}     &   43.5  \\
\textbf{COCO + COCO-A + KINS + WALT3D}             & \multicolumn{1}{c}{\textbf{38.8}}     &  \textbf{48.7}   & \multicolumn{1}{c}{\textbf{35.6}}     &   \textbf{45.3}  \\ \bottomrule
\end{tabular}%
}
\vspace{-2mm}
\caption{Finetuning with our \textbf{WALT3D} data improves generalization across KINS and COCO-A (vehicle \& people). Additionally, combining our data with domain-specific data like KINS and COCO-A further enhances performance in their domains.}
\vspace{-0.15in}
\label{tab:cocoa}
\end{table}

\noindent{\bf Cross-evaluations on pedestrian tracking:} We evaluate how our \textbf{WALT3D} data helps pedestrian tracking in Table~\ref{tab:mot}. To  highlight how better detections lead to better tracking, we use a simple tracker SORT~\cite{Bewley2016_sort} with two detectors: one pretrained on COCO and one finetuned with our \textbf{WALT3D} data. On MOT17-train, the model finetuned with \textbf{WALT3D} data produces high-quality detections under strong occlusions, resulting in improvements over the baseline in all metrics, particularly those favoring better detection quality like MOTA and DetA (see metrics details in~\cite{luiten2021hota}).

\vspace{-2mm}
\begin{table}[h]
\centering
\resizebox{0.90\columnwidth}{!}{%
\begin{tabular}{ccccc}
\toprule
\multicolumn{1}{c|}{Detector Train Data}            & MOTA$\uparrow$ & DetA$\uparrow$ & IDF1$\uparrow$ & HOTA$\uparrow$ \\ \midrule
\multicolumn{1}{l|}{COCO}                  &   42.2   &  40.3    &  46.7   &   40.1  \\
\multicolumn{1}{l|}{\textbf{COCO + WALT3D}} &  \textbf{47.2}    &   \textbf{43.2}   &  \textbf{49.5}   &   \textbf{42.0}   \\ \bottomrule
\end{tabular}%
}
\vspace{-2mm}
\caption{MOT17-train results using SORT with two  detectors: one pretrained on COCO and one fine-tuned with our WALT3D data.}
\label{tab:mot}
\end{table}
\vspace{-0.1in}

%

\begin{figure}[t!]
    \centering
\begin{sideways}\ \ \ \ \scriptsize{WALT}\end{sideways}
        \includegraphics[width=0.10\textwidth]{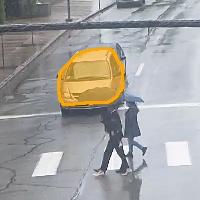}
        \includegraphics[width=0.10\textwidth]{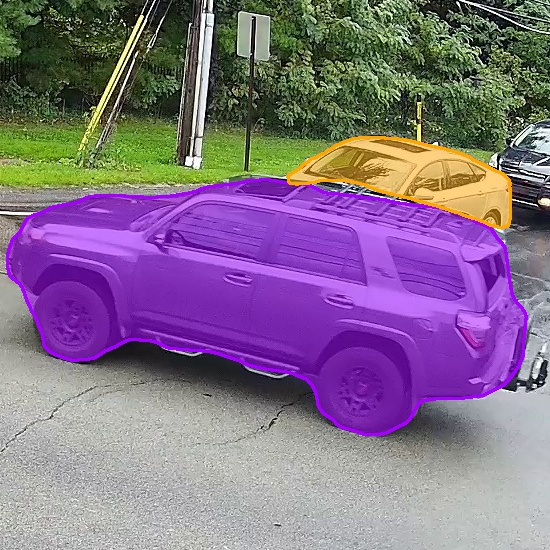}
        ~~\rulesep~~
        \includegraphics[width=0.10\textwidth]{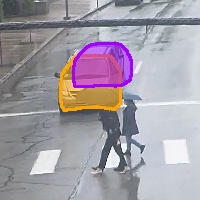}
        \includegraphics[width=0.10\textwidth]{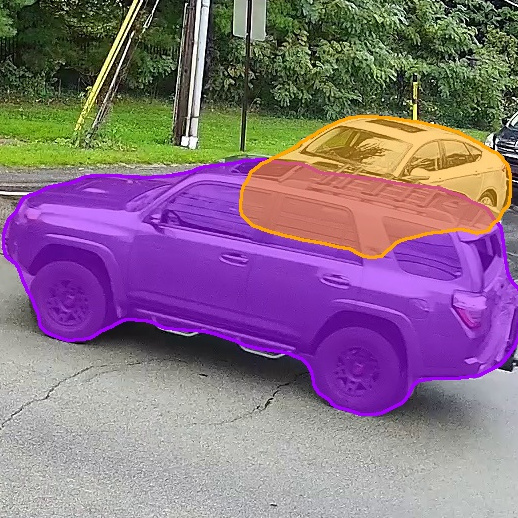}

\begin{sideways}\ \ \scriptsize{OCC-Net}\end{sideways}
        \includegraphics[width=0.10\textwidth]{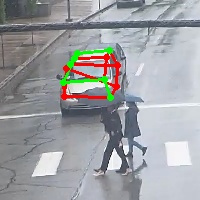}
        \includegraphics[width=0.10\textwidth]{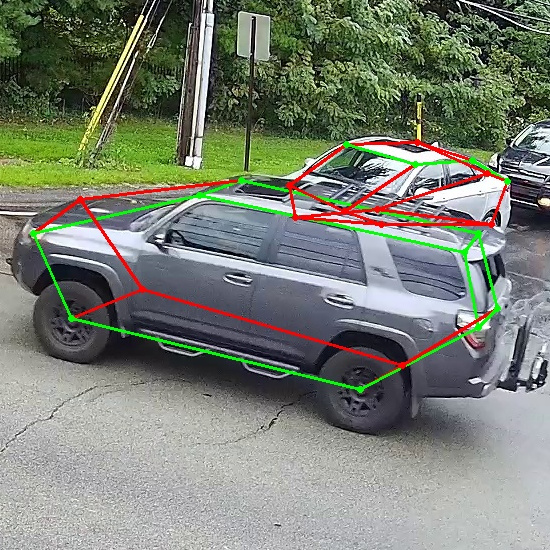}
        ~~\rulesep~~
        \includegraphics[width=0.10\textwidth]{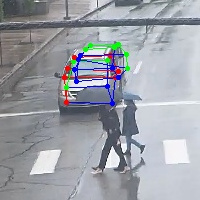}
        \includegraphics[width=0.10\textwidth]{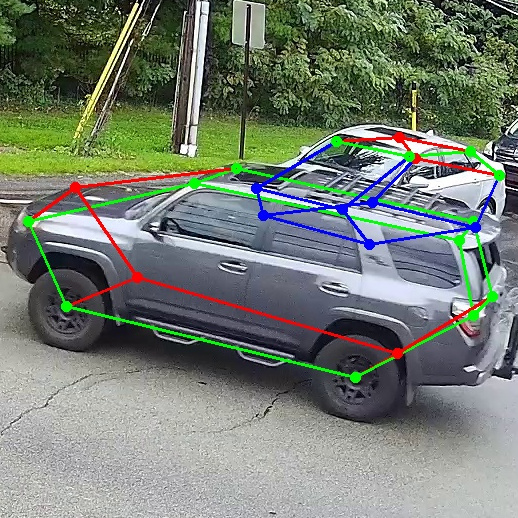}

\begin{sideways}\  \scriptsize{3DRCNN}\end{sideways}
        \includegraphics[width=0.10\textwidth]{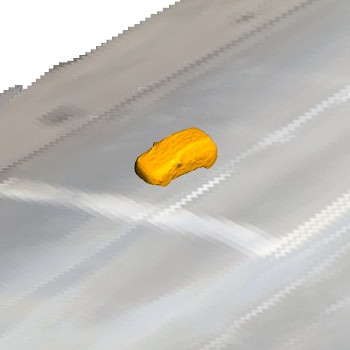}
        \includegraphics[width=0.10\textwidth]{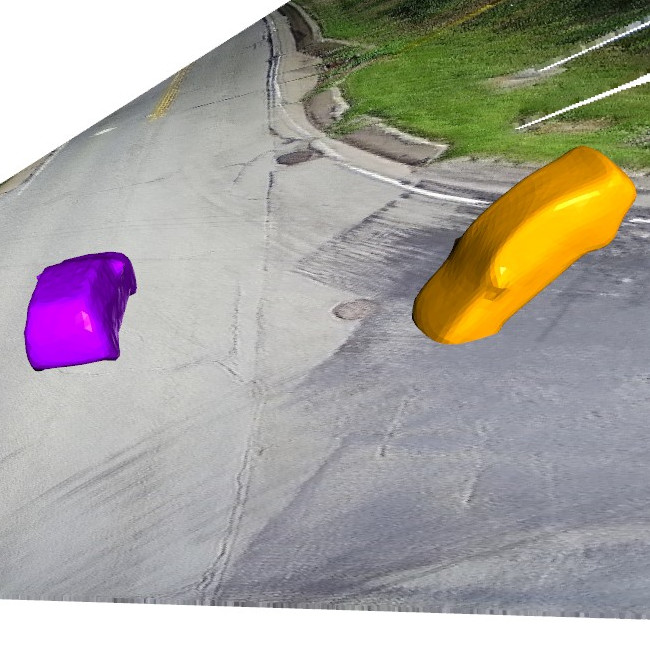}
        ~~\rulesep~~
        \includegraphics[width=0.10\textwidth]{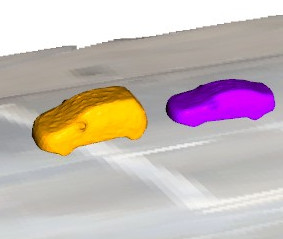}
        \includegraphics[width=0.10\textwidth]{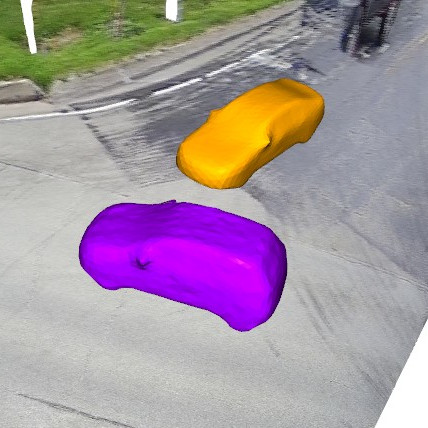}
        Previous SOTA \ \ \ \ \ \ \ \ \ \ \ \ \ \ \ \ \ \ \ \ \ \ \ \ \ \ \ Ours  

\vspace{-0.1in}
\caption{Qualitative results showing that our data improves amodal segmentation, keypoint prediction and 3D reconstruction compared to previous SOTA.}
\label{fig:qual}
 \vspace{-0.2in}
\end{figure}

\begin{figure*}[t!]
    \centering
    \includegraphics[width=\textwidth]{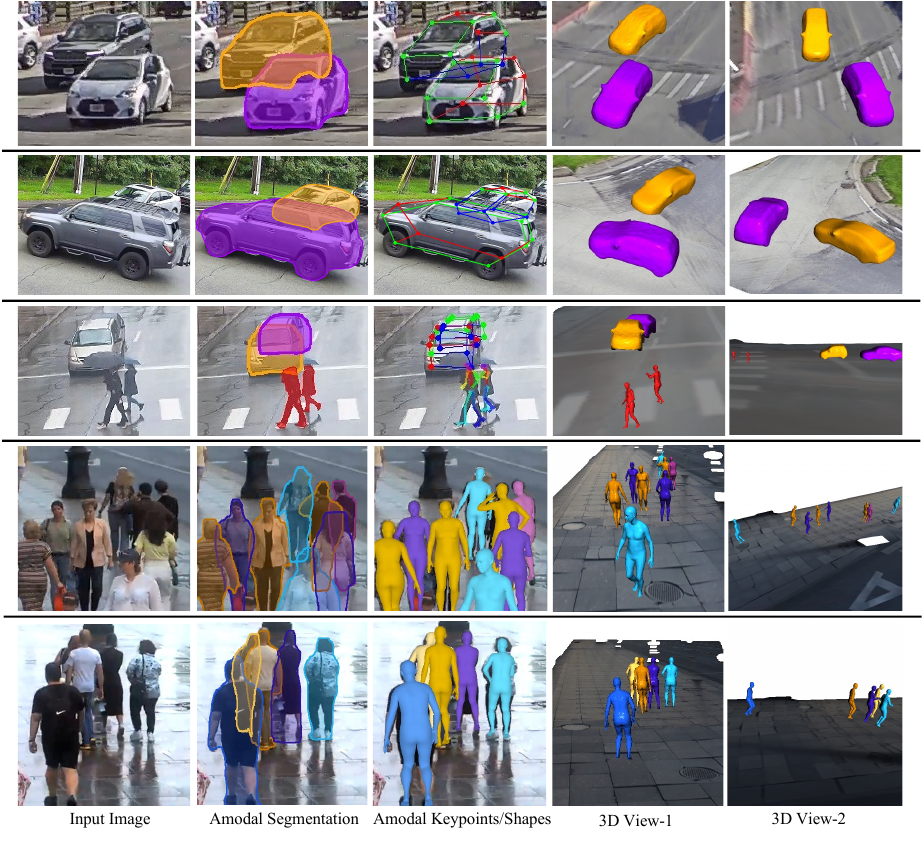}
    \vspace{-0.35in}
    \caption{We show qualitative results of our method on real data. Our method produces accurate amodal segmentation, keypoints, as well as 3D poses and shapes across diverse poses and occlusion configurations. In particular, we show results on different level and types of occlusions like vehicle-vehicle \textbf{(row 1, 2, 3)}, vehicle-people \textbf{(row 3)}, and people-people occlusion \textbf{(row 4, 5)}.}
    \label{fig:final_results}
    \vspace{-0.2in}
\end{figure*}

\noindent \textbf{Qualitative Results:} Results of our method can be seen in Fig.~\ref{fig:qual} and Fig.~\ref{fig:final_results}. Comparing to previous methods, observe that our method can produce robust 2D/3D predictions, even under challenging high-occlusion scenarios.
\section{Discussion and Conclusion}
\vspace{-0.05in}
Our data generation framework is method-agnostic, accommodating various robust object pose estimation alternatives~\cite{wang2021nemo, ma2022robust} as well as methods like VIBE~\cite{kocabas2020vibe} for high-quality human reconstructions from videos.
Moreover, as our pseudo-groundtruth data includes metric-scale depth information using ground plane, we can augment existing datasets like Relative Human~\cite{Sun_2022_CVPR} to enhance the robustness of 3D multi-human reconstruction methods to occlusion.

\noindent \textbf{Limitations:} Further research is required for generalization across diverse views for it to be used as a generic solution. It also assumes a mean shape or a parametric object model is available, posing challenges for rare objects. Addressing appearance inconsistencies (e.g., variations in lighting) in clip-art images is a promising research direction.

In conclusion, our method introduces an automated approach to generate a realistic dataset for reconstructing dynamic objects under occlusions from time-lapse imagery. It demonstrates significant improvements in both 2D and 3D object reconstruction, particularly in busy urban scenes with diverse occlusion configurations.

\noindent \textbf{Potential Societal Impact.} We do not perform any human subjects research from these cameras. For privacy, we blur faces and license plates in all images intended for release.

\small \noindent \textbf{Acknowledgements:} This work was supported in part by an NSF Grant CNS-2038612, a US DOT grant 69A3551747111 through the Mobility21 UTC and grants 69A3552344811 and 69A3552348316 through the Safety21 UTC.\clearpage
{
    \small
    \bibliographystyle{ieeenat_fullname}
    \bibliography{main}
}
\clearpage

\renewcommand\thesubsection{\Alph{subsection}}

\subsection{Summary and More Results}

\textcolor{red}{For a brief summary of our method and additional results, we highly encourage the readers to check out the included short video.}


\subsection{Vehicle 2D Keypoints Dataset}
As mentioned in the main paper, although existing datasets like PASCAL3D+~\cite{Xiang_Pascal}, KITTI-3D~\cite{li2017deep}, Carfusion~\cite{Reddy_2018_CVPR}, and ApolloCar3D~\cite{apo} provide annotated 2D vehicle keypoints, they mostly focus on driving scenes~\cite{Reddy_2018_CVPR, li2017deep, apo} or have limited training examples~\cite{Xiang_Pascal}, lacking the necessary appearance diversity. To increase the dataset diversity, we prioritized the number of different cameras and viewpoints rather than the number of images per camera. A summary and comparison of our proposed Vehicle 2D Keypoints dataset with other publicly available datasets are detailed in Table~\ref{tab:dataset_supp}. On average, we extracted 120 images per camera source for more than 60 different cameras spanning a wide variety of viewpoints, appearances, sensor types, etc. For each image, we run an off-the-shelf object detector to extract the car instances with high confidence score. This set of car instances are manually annotated by the trained annotators from a commercial annotation service. We utilized a web-based interface annotation tool from DeepLabCut~\cite{Mathisetal2018} where the annotators were asked to select 12 keypoint locations and its corresponding occlusion category (visible/self-occluded/occluded-by-others) for every car. Note that we also asked the annotators to filter out erroneous instances such as bad quality images and/or wrong detections. As of the time of paper submission, we have annotated a total of 42,547 car instances in 7,018 images. 

\subsection{Camera Intrinsics and Ground Plane} 
We follow Vuong et al.~\cite{vuong2024trafficcalib} to obtain the intrinsic parameters and ground plane equation for each of the stationary traffic camera. Specifically, we used the panorama images from Google Street View (GSV)~\cite{GoogleStreetView} to build a metric 3D scene reconstruction (at the desired camera location), then the stationary camera is registered within the reconstruction to determine its intrinsic and extrinsic parameters.
We also geo-register the scene to a metric scale using the GPS coordinates, and the road plane equation is estimated by fitting a plane to the set of 3D points whose 2D pixel projections belong to the \textit{road} category obtained from off-the-shelf semantic segmentation method~\cite{cheng2021mask2former}. 
The camera poses and plane equation are used in 3D reconstruction pipeline to reconstruct unoccluded objects as described in the main paper.
Thanks to the ground plane geometry constraint, we can reconstruct the accurate 3D geometry of cars and pedestrians, generating realistic occlusion configurations.
This method enables us to obtain accurate calibration for more than 100 stationary cameras worldwide, thus allowing for a significant expansion of our clip-art dataset.

\subsection{Benchmarking on Additional Datasets}

\begin{table}[h!]
\centering
\begin{adjustbox}{width=0.45\textwidth}

{\footnotesize
\centering
 \setlength{\tabcolsep}{2.5pt}
 \begin{tabular}{ccccccc}
 \toprule
 \centering
    $Acc-\frac{\pi}{6}$ & i.i.d & shape & pose & texture & context & weather\\
    \midrule
NeMo & 66.7 & \textbf{51.7} & \textbf{56.9} & 52.6 & \textbf{51.3} & 49.8 \\
Ours & \textbf{75.4} & 48.6 & 50.8 & \textbf{56.7} & 49.1 & \textbf{55.6} \\
\bottomrule
 \end{tabular}
 
 }
\end{adjustbox}
 \caption{\footnotesize Comparisons on the OOD-CV~\cite{zhao2022ood} dataset (car).}
 \label{tab:ood-cv}
\end{table}

\begin{figure}[t!]
    \centering
        \includegraphics[width=0.10\textwidth, height=0.09\textwidth]{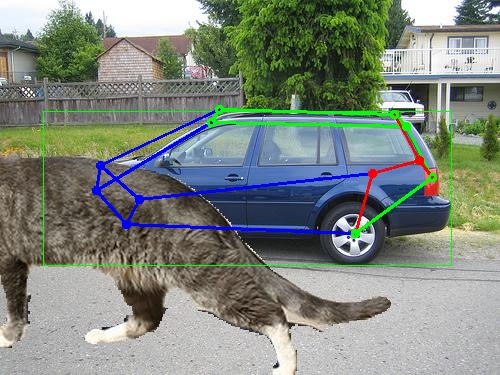}
        \includegraphics[width=0.11\textwidth, height=0.09\textwidth]{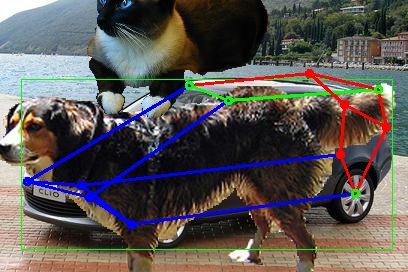}
       ~~\rulesep~~
        \includegraphics[width=0.10\textwidth, height=0.09\textwidth]{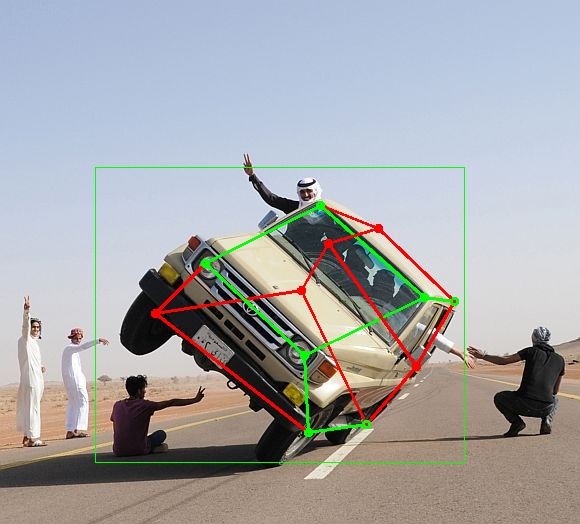}
        \includegraphics[width=0.11\textwidth, height=0.09\textwidth]{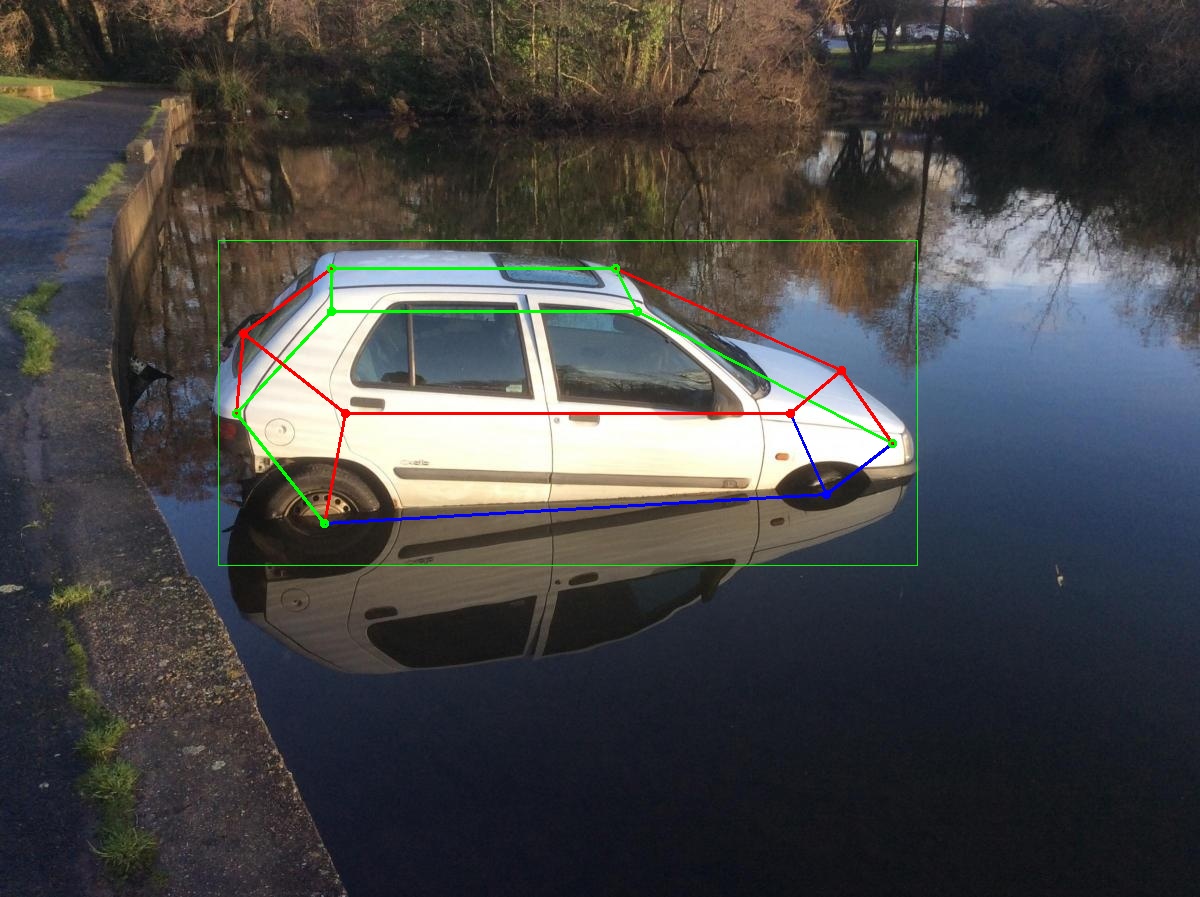}

\caption{\footnotesize Qualitative Results on OccludedPascal3D+ (left) and OOD-CV (right) dataset.}
\label{fig:ood-cv}
\end{figure}

\noindent \textbf{Evaluation on OccludedPascal3D+ dataset:} Table~\ref{tab:compare} shows that our method performs better than NeMo~\cite{wang2021nemo} and Ma et al.~\cite{ma2022robust} on the OccludedPascal3D+~\cite{wang2021nemo} dataset.   

\begin{table}[h!]
\begin{adjustbox}{width=0.49\textwidth}
{\footnotesize
\centering
 \setlength{\tabcolsep}{2.5pt}
 \begin{tabular}{c|ccc|ccc|ccc|ccc}
 \toprule
 \centering

   Method & \multicolumn{3}{c|}{$Acc \ (\frac{\pi}{6})$} & \multicolumn{3}{c|}{$Acc \ (\frac{\pi}{18})$} & \multicolumn{3}{c|}{Med Pose Err} & \multicolumn{3}{c}{Med ADD}  \\
   \cline{2-13}
  & L1 & L2 & L3 & L1 & L2 & L3 & L1 & L2 & L3 & L1 & L2 & L3 \\
 \midrule
 \multicolumn{13}{c}{Occluded PASCAL3D+ Dataset (car)} \\ 
 \midrule
 NeMo & 48.3 & 34.3 & 18.2 & 17.4 & 9.6 & 3.3 & 0.5 & 1.0 & 2.4 & 1.9 & 2.0 & 2.3 \\
 Ma et al. & 66.6 & 47.9 & 27.4 & 30.8 & 16.2 & 5.3 & 0.3 & 0.6 & 1.1 & 0.8 & 1.2 & 1.9  \\
 Ours & \textbf{70.4} & \textbf{56.5} & \textbf{35.3} & \textbf{36.8} & \textbf{25.4} & \textbf{15.3}  & \textbf{0.2} & \textbf{0.4} & \textbf{0.8} & \textbf{0.6} & \textbf{1.0} & \textbf{1.4}  \\
 \bottomrule
 \end{tabular}
 
 }
\end{adjustbox}
 \caption{\footnotesize  Baseline comparisons across object pose metrics on Occluded-PASDAL3D+~\cite{wang2021nemo} for vehicle category.}
 \label{tab:compare}
\end{table}

\noindent \textbf{Evaluation on OOD-CV dataset:} Quantitative results on OOD-CV~\cite{zhao2022ood} dataset are shown in Table~\ref{tab:ood-cv}. Although our method has never been trained on the anomalous scenarios in this dataset, our approach shows higher performances on many testing subsets. Please see qualitative results in Fig.~\ref{fig:ood-cv}.

\begin{table}[h]
\centering
\resizebox{0.9\columnwidth}{!}{
\begin{tabular}{l|c|c|c|c|c}
\hline
Metric    & $\delta=0.01$ & $\delta=0.1$ & $\delta=0.2$ & $\delta=0.5$ & \textbf{OC (ours)} \\ \hline
Recall    & 0.60     & 0.42    & 0.17    & 0.01    & \textbf{0.81}      \\ 
Precision & 0.32     & 0.41    & 0.52    & 0.57    & \textbf{0.70}      \\ \hline
\end{tabular}}
\vspace{-0.1in}
\caption{Accuracy of our OC module compared with  baseline using bbox IOU threshold $\delta$ in detecting unoccluded objects.}
\label{table:occ_accuracy}
\vspace{-0.1in}
\end{table}

\noindent \textbf{Mining Unoccluded Objects:}
To identify unoccluded objects, we evaluate two methods: a simple heuristic based on bounding box IOU threshold $\delta$ (as used in WALT~\cite{Reddy_2022_CVPR}) and training an Occlusion Classifier (OC) using human-annotated data (using images from our new vehicle keypoints dataset). 
Table~\ref{table:occ_accuracy} demonstrates that our OC module is more effective than the heuristic, particularly in inter-category occlusion scenarios (e.g., vehicles occluded by people or background objects). 
This allows us to efficiently filter out unwanted occluded objects in the training data, improving data purity.
While not essential for our method, we believe this human-annotated dataset is important for future research on understanding and handling occlusion.

\begin{table*}[ht!]
\resizebox{\textwidth}{!}{
\begin{tabular}{c|c|cccc|c|c|c|c}
\hline
\multirow{2}{*}{Dataset} &
  \multirow{2}{*}{Image source} &
  \multicolumn{4}{c|}{Appearance diversity in terms of} &
  \multirow{2}{*}{\# images} &
  \multirow{2}{*}{\# car instances} &
  \multirow{2}{*}{Occ. keypoint annotations} &
  \multirow{2}{*}{Per-keypoint occ. type} \\ \cline{3-6}
 &
   &
  \multicolumn{1}{c|}{Cities} &
  \multicolumn{1}{c|}{Times of Day} &
  \multicolumn{1}{c|}{Weathers} &
  Viewpoints &
   &
   &
   &
   \\ \hline
PASCAL3D+   & Natural      & \multicolumn{1}{c|}{Yes} & \multicolumn{1}{c|}{Yes} & \multicolumn{1}{c|}{Yes} & No & 6,704  & 7,791   & No  & No \\ 
KITTI-3D    & Self-driving & \multicolumn{1}{c|}{No}  & \multicolumn{1}{c|}{No}  & \multicolumn{1}{c|}{No}  & No & 2,040  & 2,040   & No  & No \\ 
Carfusion   & Handheld     & \multicolumn{1}{c|}{No}  & \multicolumn{1}{c|}{No}  & \multicolumn{1}{c|}{No}  & No & 53,000 & 100,000 & Yes & No \\ 
ApolloCar3D & Self-driving & \multicolumn{1}{c|}{No}  & \multicolumn{1}{c|}{No}  & \multicolumn{1}{c|}{No}  & No & 5,277  & 60,000  & No  & No \\ \hline
\textbf{Ours} &
  \begin{tabular}[c]{@{}c@{}}Handheld\\ Self-driving\\ Traffic cameras\end{tabular} &
  \multicolumn{1}{c|}{Yes} &
  \multicolumn{1}{c|}{Yes} &
  \multicolumn{1}{c|}{Yes} &
  Yes &
  7,018 &
  42,547 &
  Yes &
  Yes \\ \hline
\end{tabular}}
\caption{Summary and comparison of our \textbf{Vehicle 2D Keypoints dataset} to other publicly available datasets.}
\label{tab:dataset_supp}
\end{table*}

\subsection{Additional 2D/3D Clip-Art Data Examples} 

\begin{figure*}[h!]
    \centering

        \fbox{\includegraphics[width=0.232\textwidth,height=0.15\textwidth]{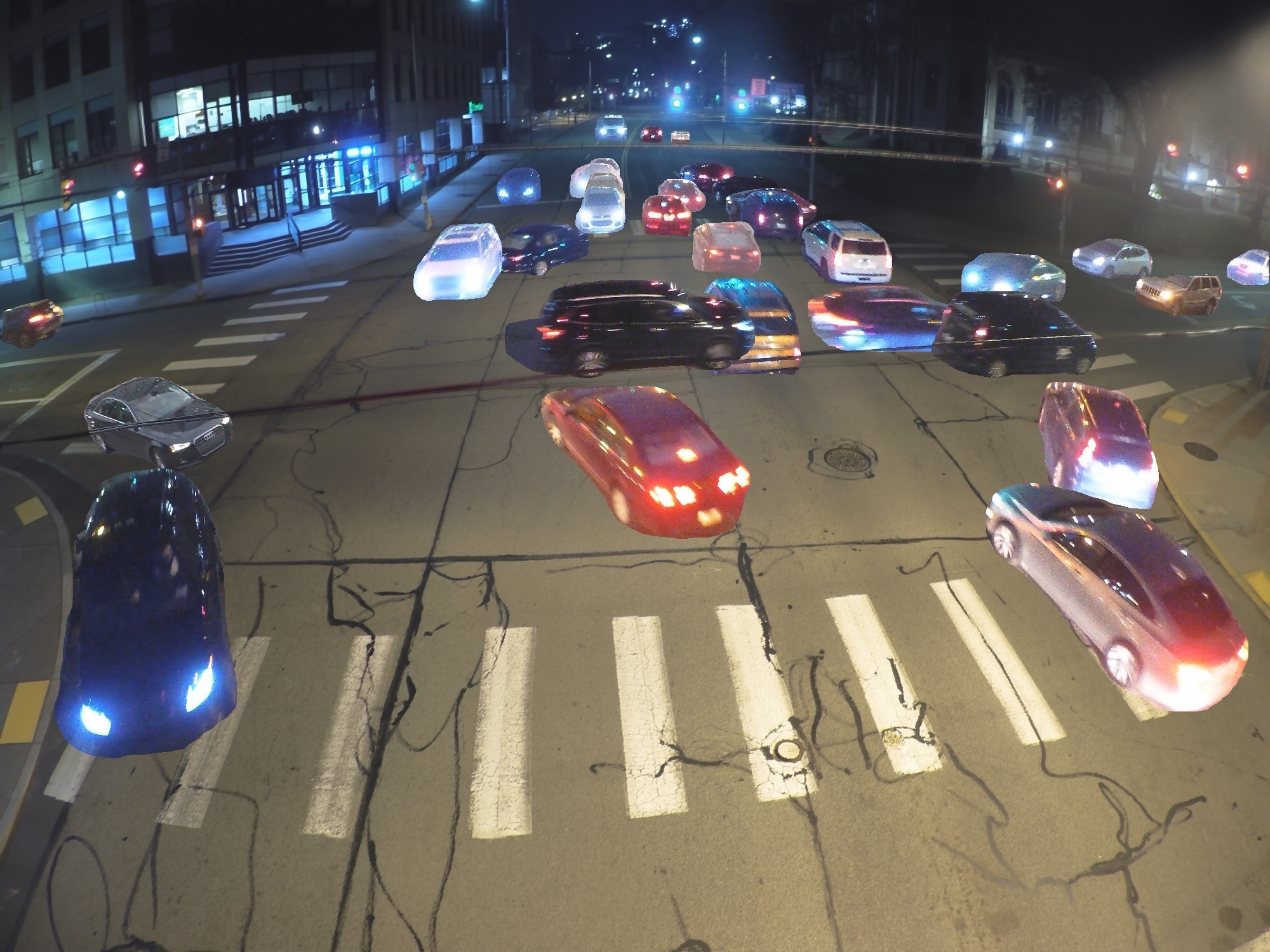}}
        \fbox{\includegraphics[width=0.232\textwidth,height=0.15\textwidth]{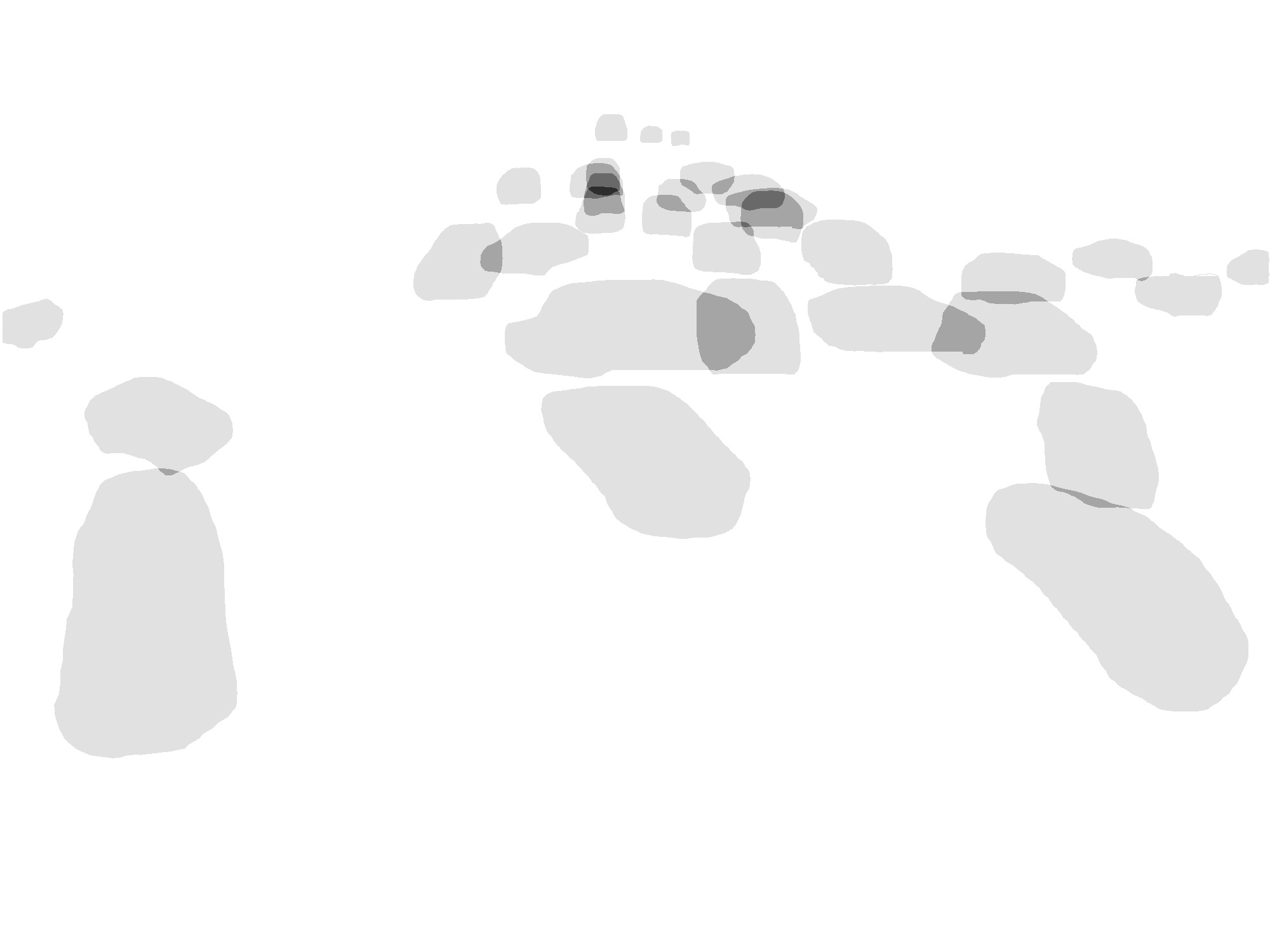}}
        \fbox{\includegraphics[width=0.232\textwidth,height=0.15\textwidth]{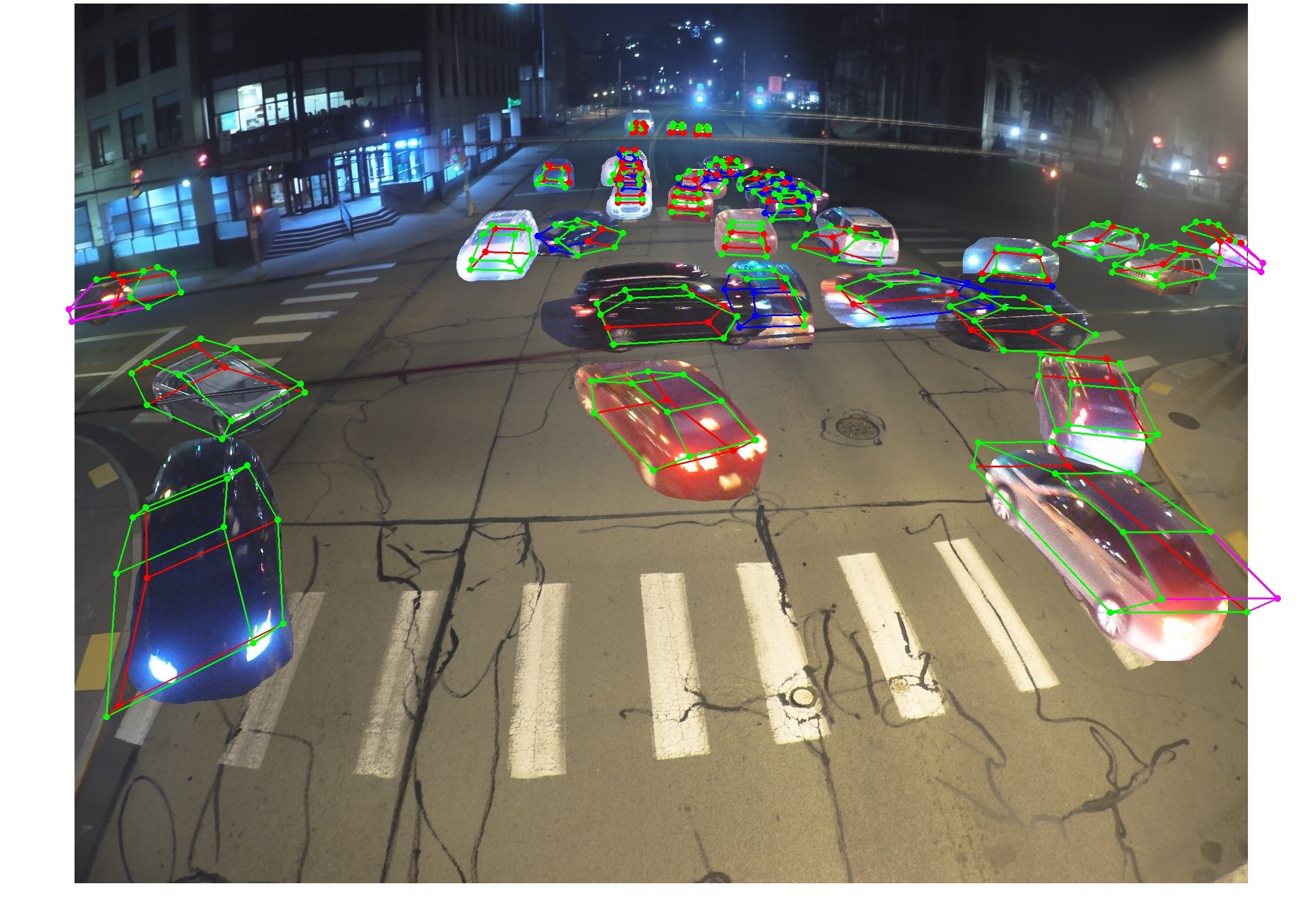}}
        \fbox{\includegraphics[width=0.232\textwidth,height=0.15\textwidth]{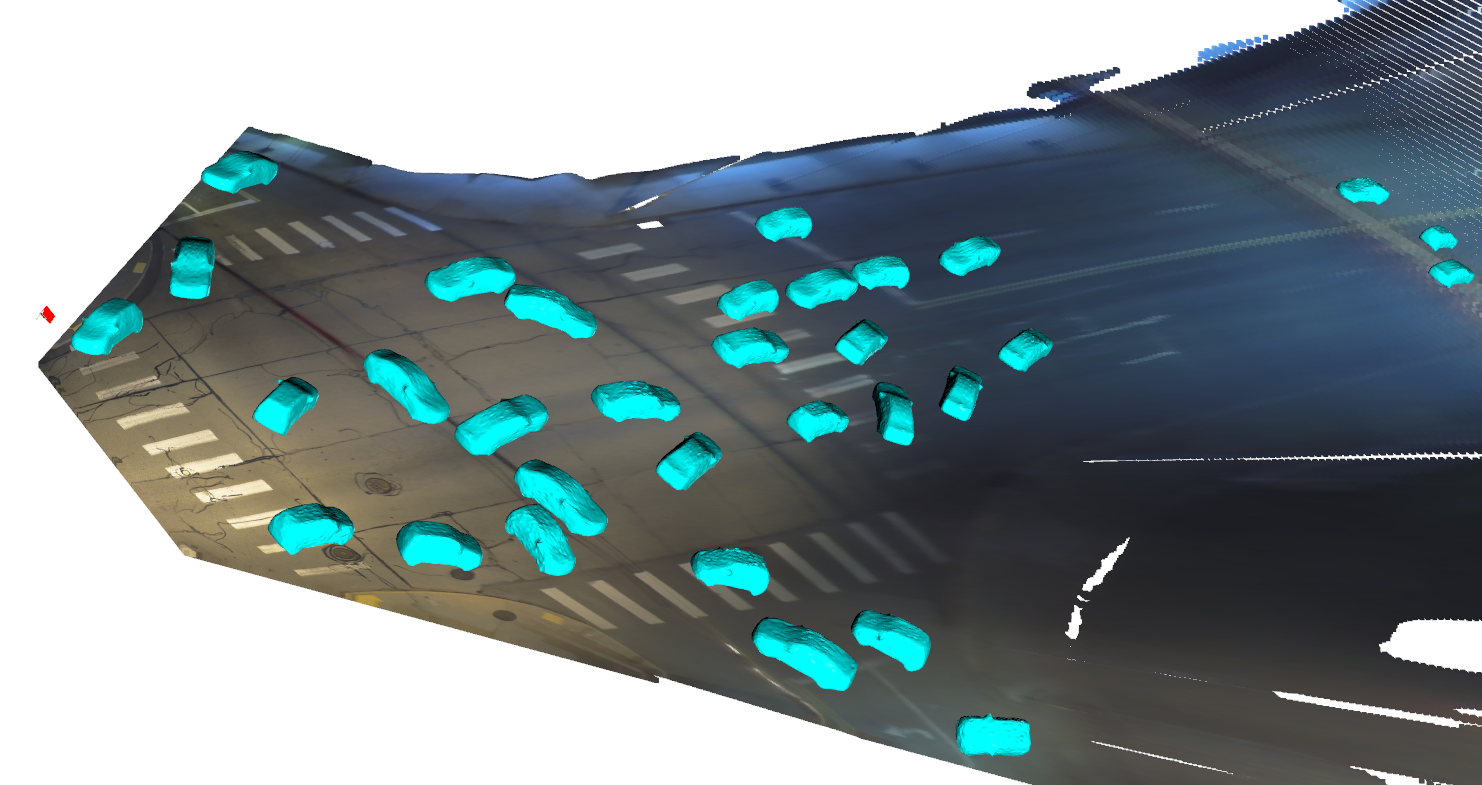}}

        \fbox{\includegraphics[width=0.232\textwidth,height=0.15\textwidth]{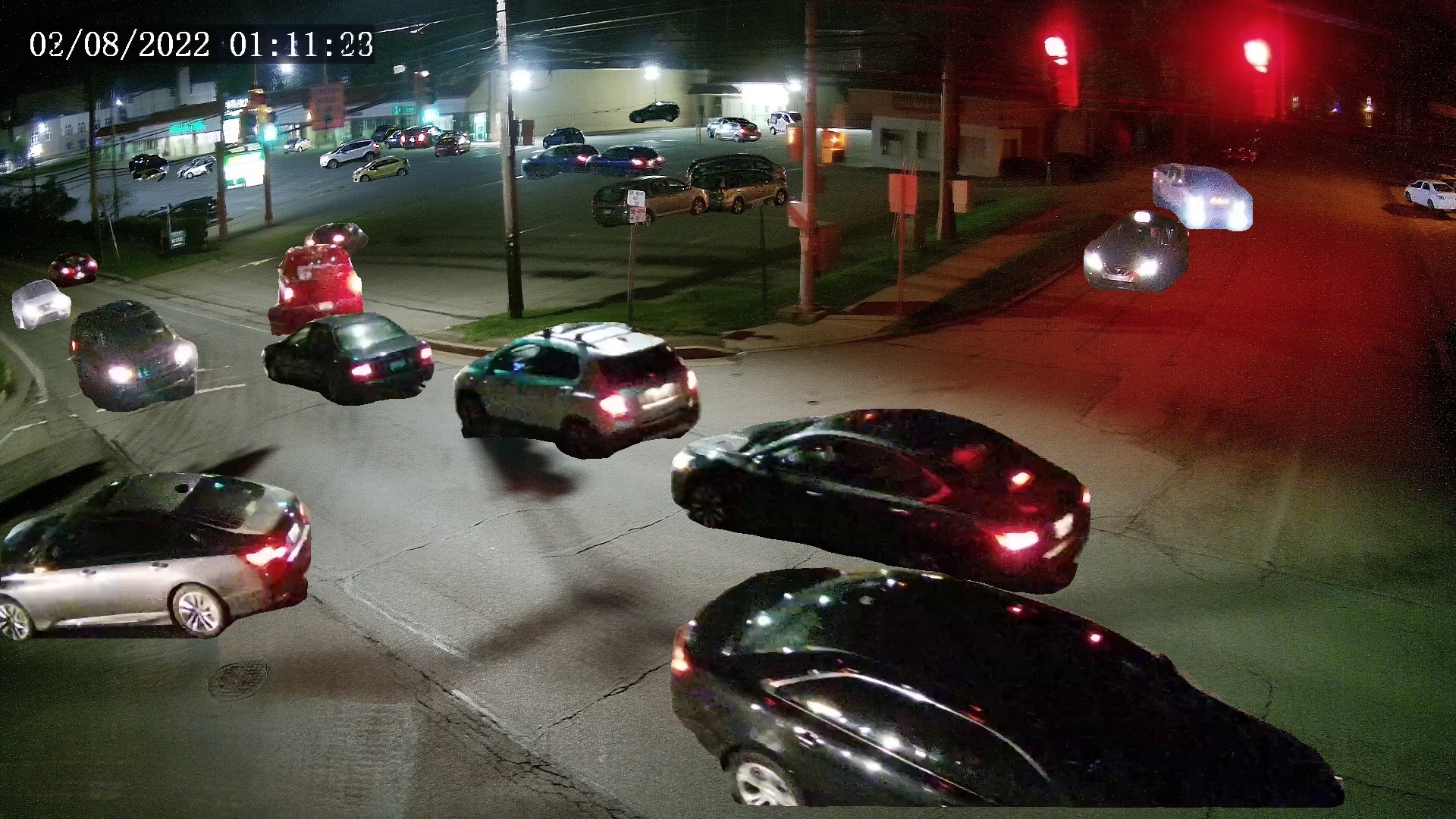}}
        \fbox{\includegraphics[width=0.232\textwidth,height=0.15\textwidth]{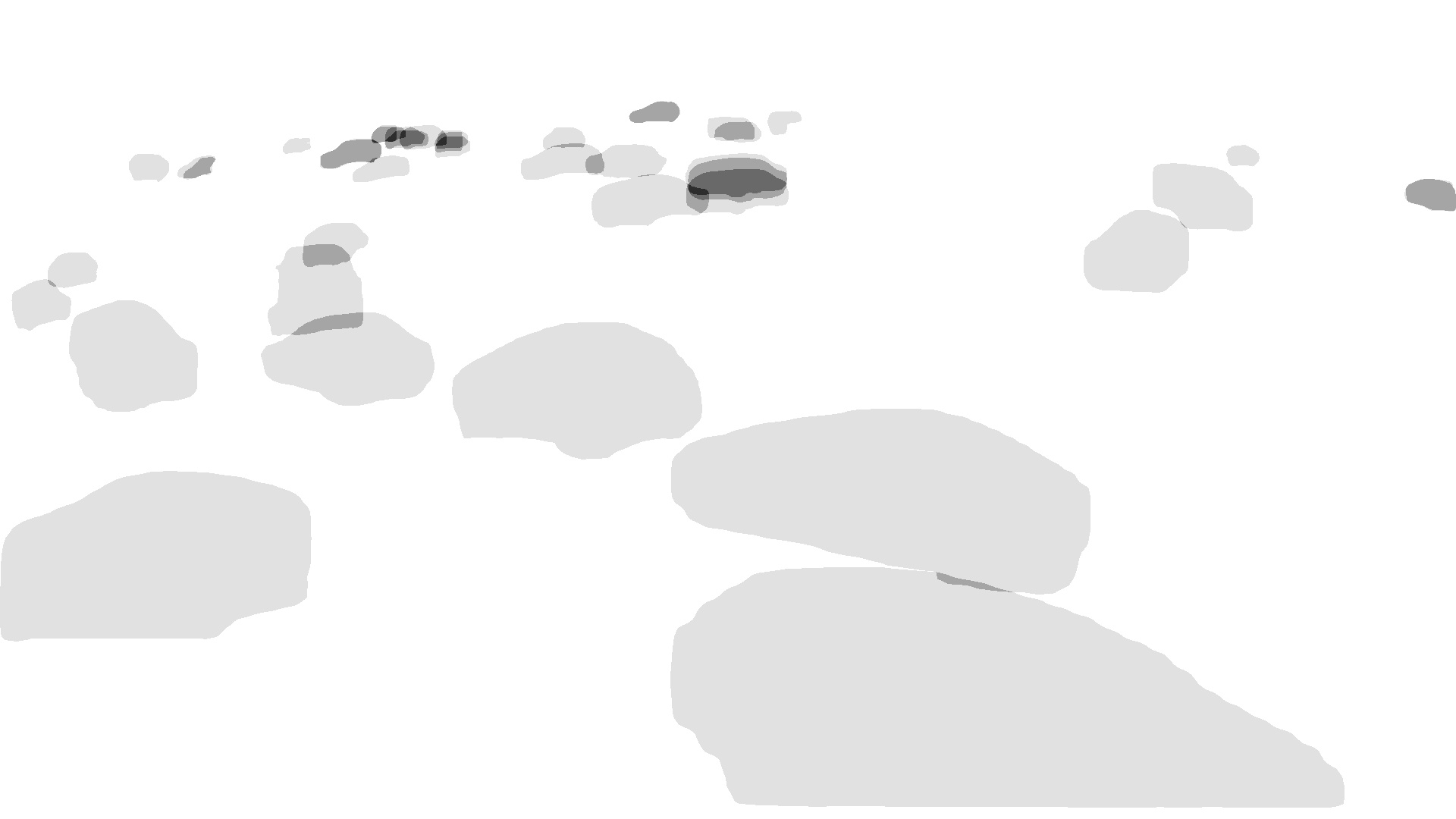}}
        \fbox{\includegraphics[width=0.232\textwidth,height=0.15\textwidth]{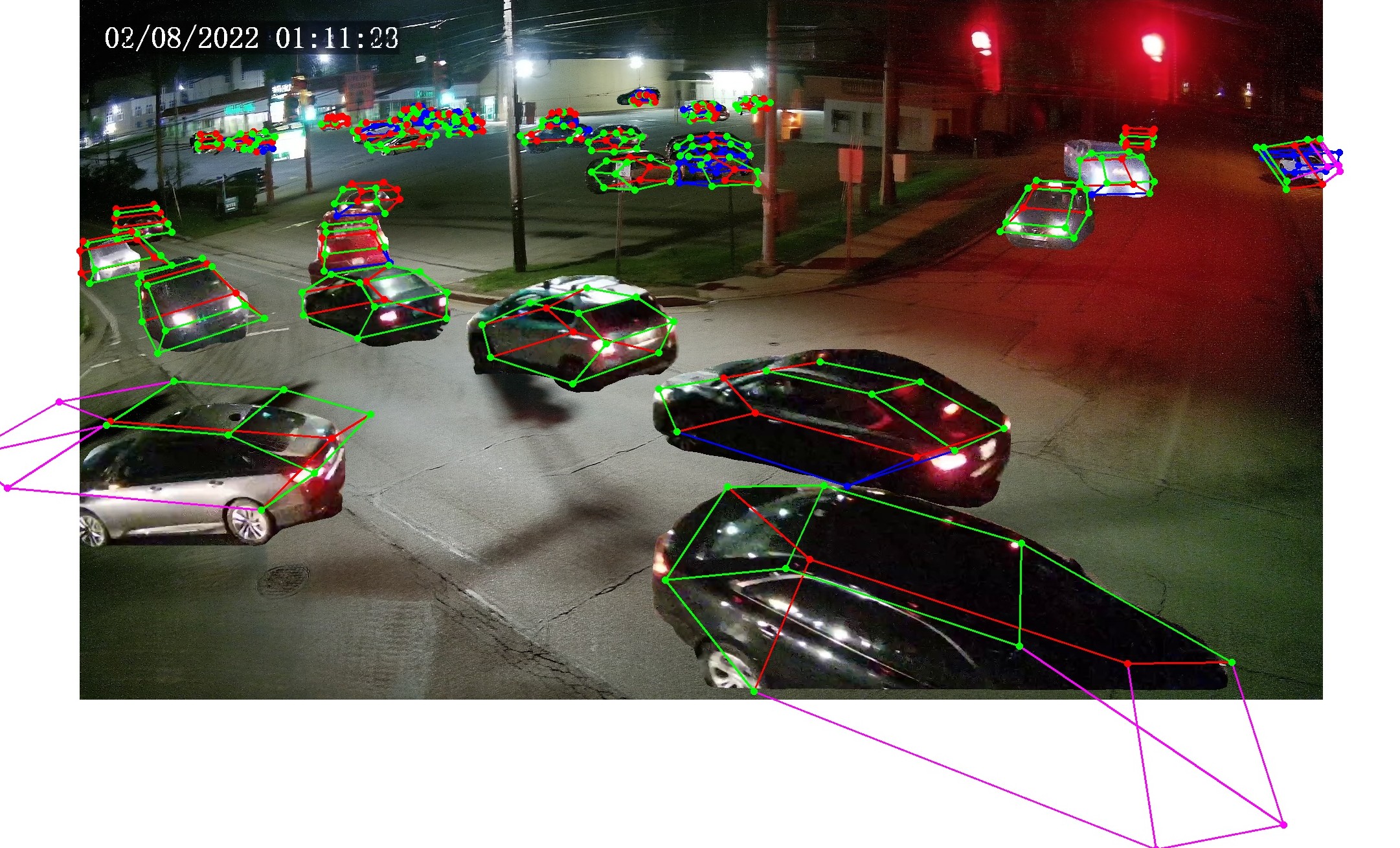}}
        \fbox{\includegraphics[width=0.232\textwidth,height=0.15\textwidth]{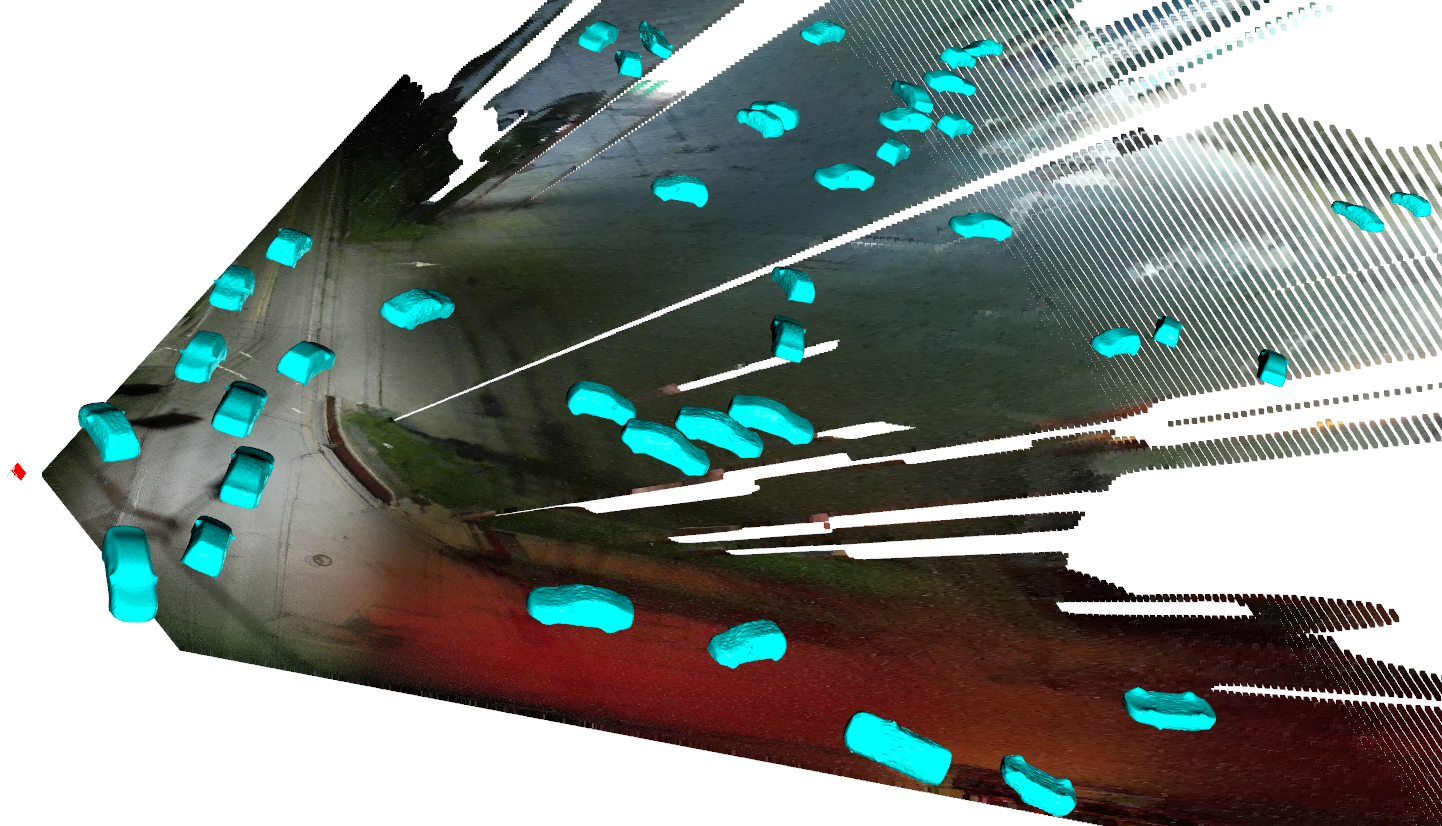}}

         \fbox{\includegraphics[width=0.232\textwidth,height=0.15\textwidth]{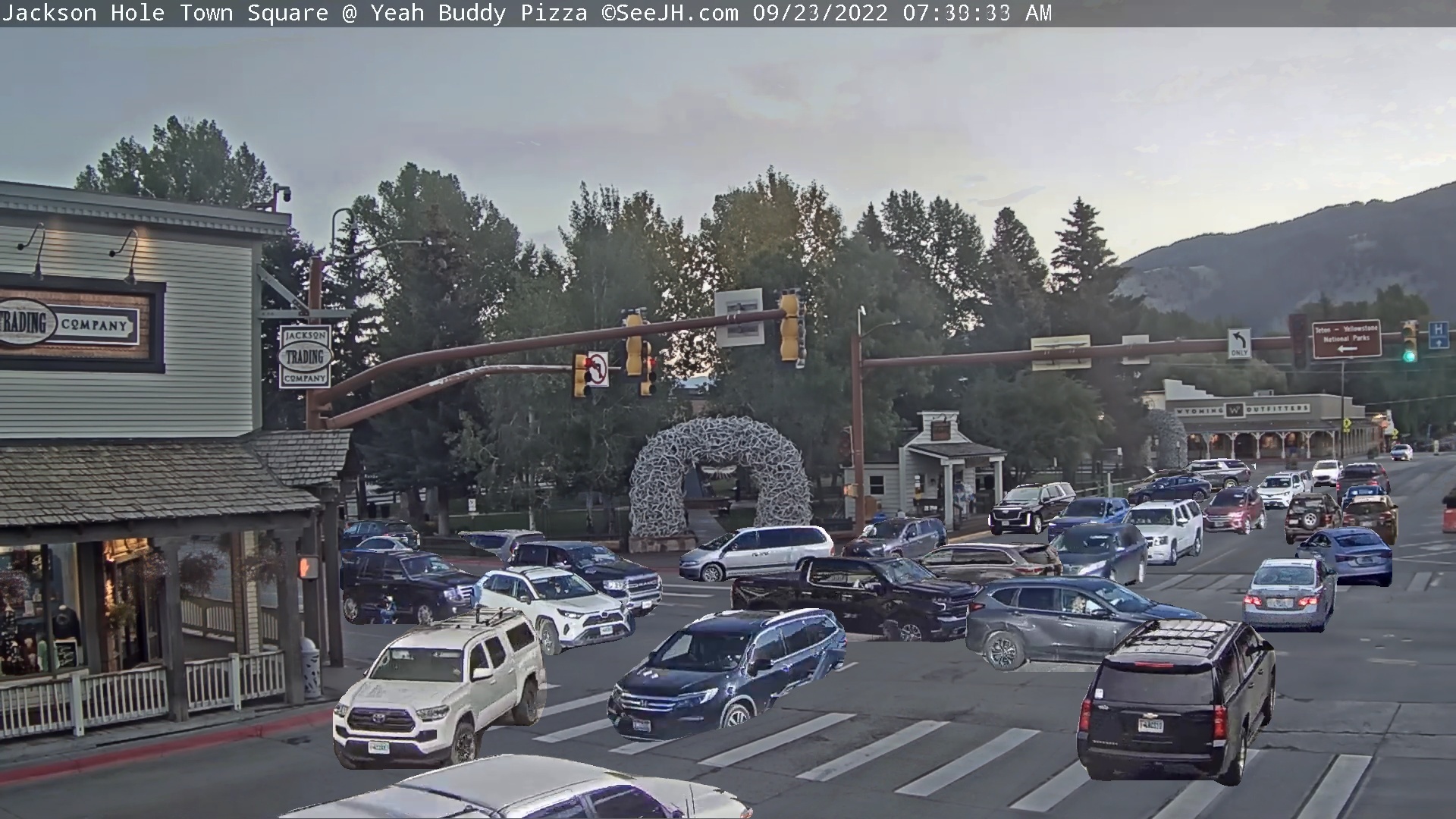}}
         \fbox{\includegraphics[width=0.232\textwidth,height=0.15\textwidth]{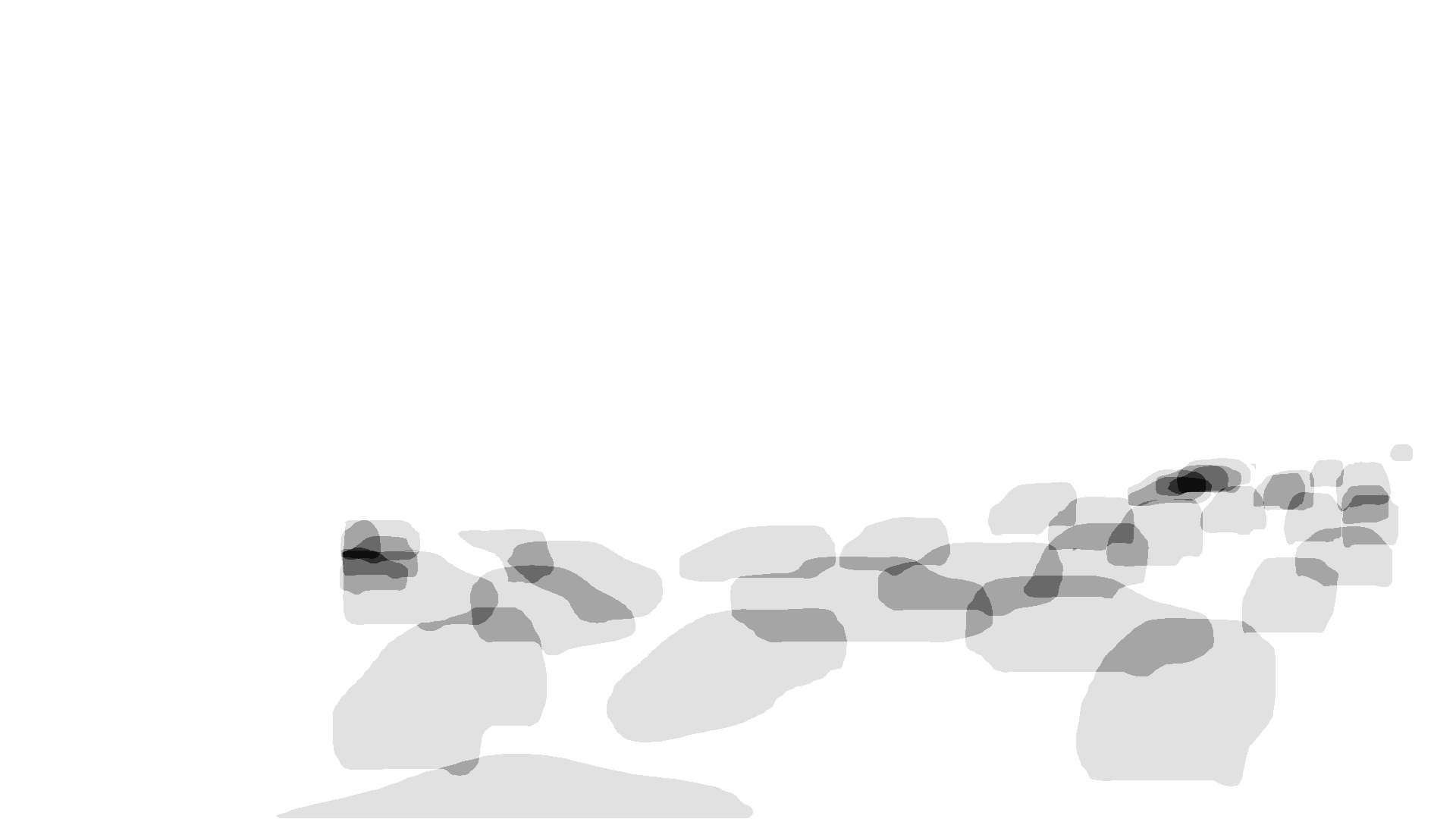}}
         \fbox{\includegraphics[width=0.232\textwidth,height=0.15\textwidth]{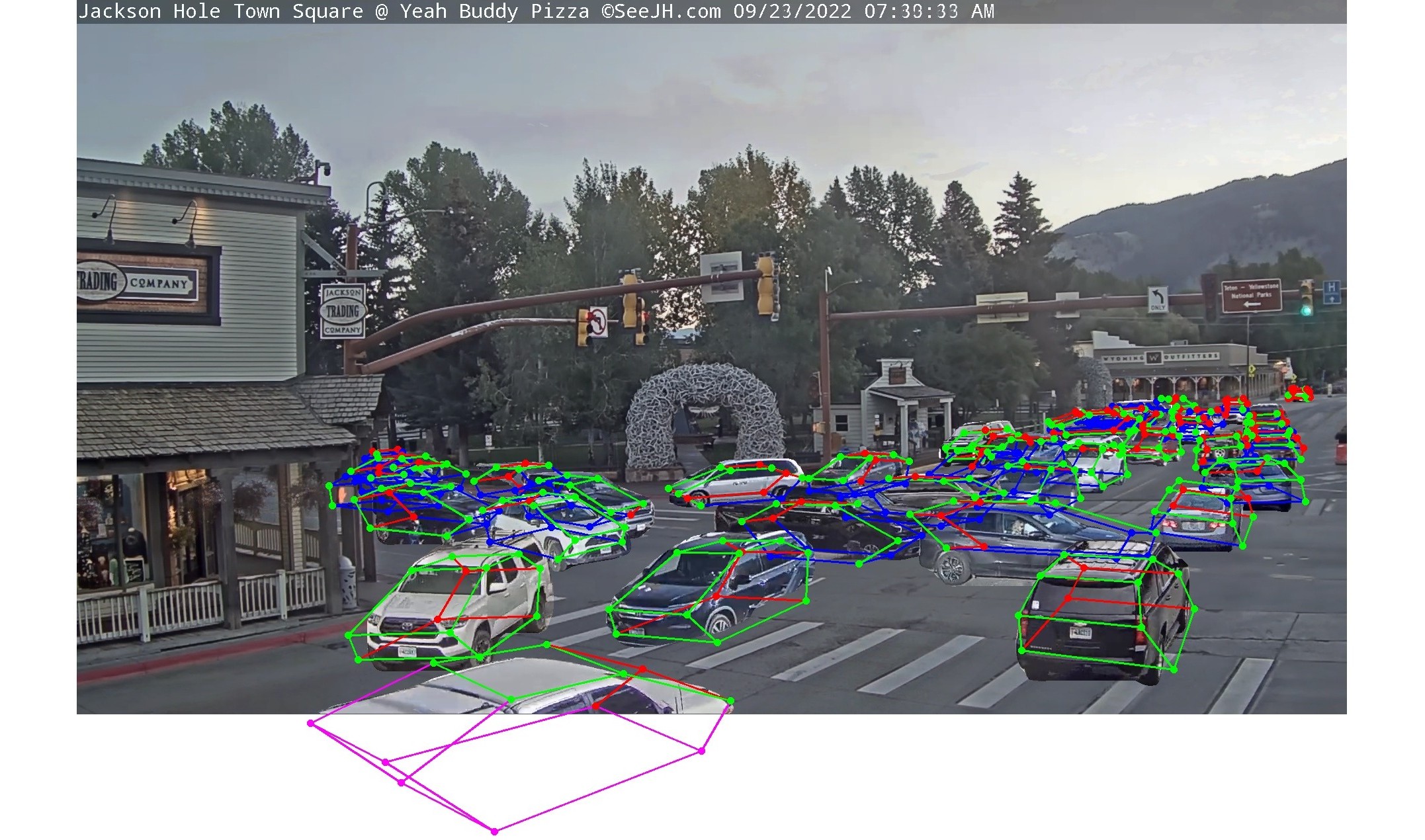}}
         \fbox{\includegraphics[width=0.232\textwidth,height=0.15\textwidth]{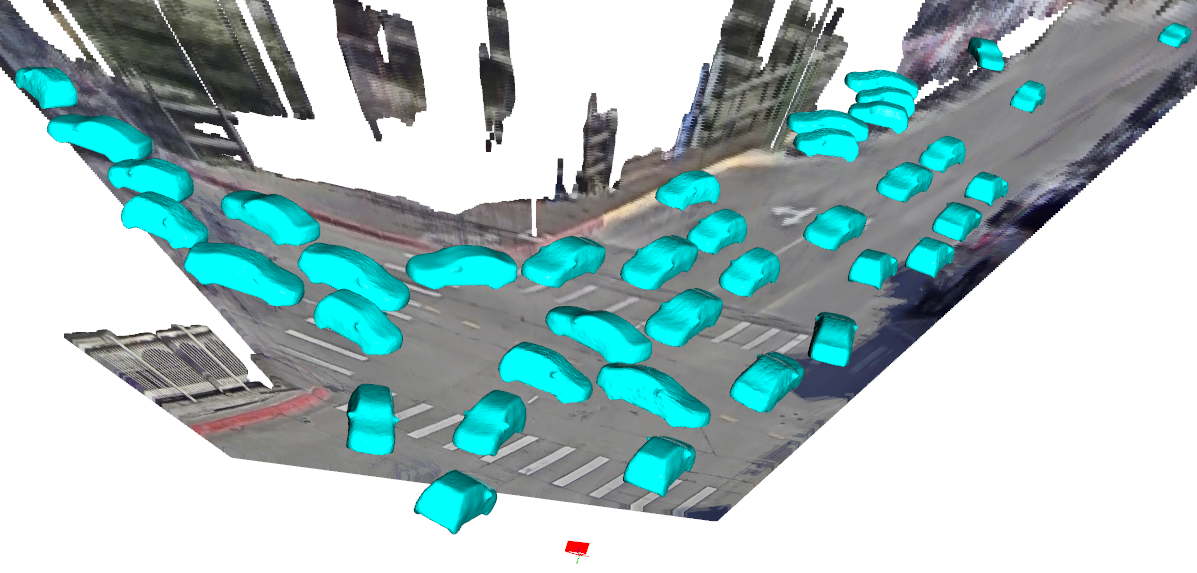}}
        
         \fbox{\includegraphics[width=0.232\textwidth,height=0.15\textwidth]{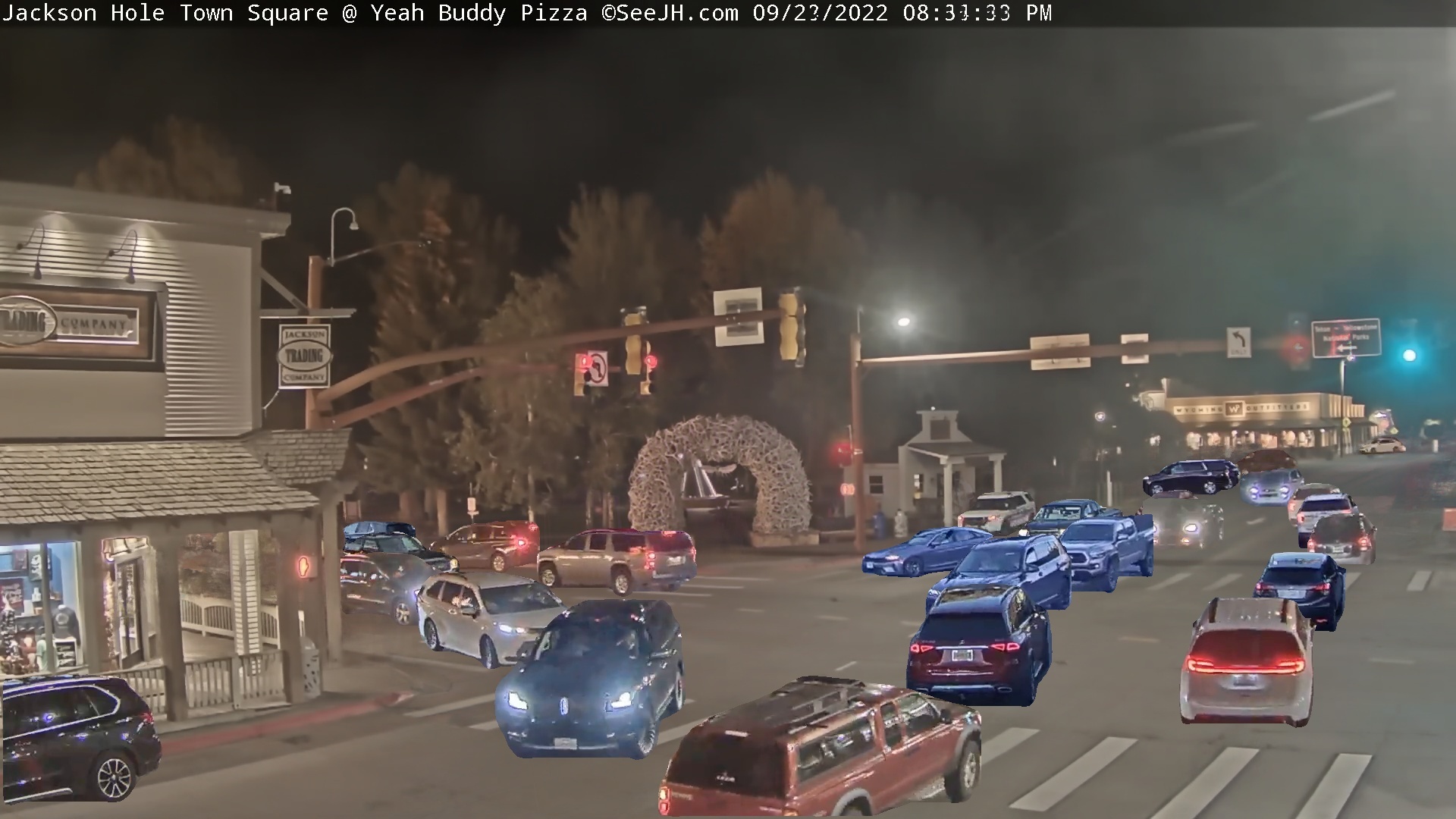}}
         \fbox{\includegraphics[width=0.232\textwidth,height=0.15\textwidth]{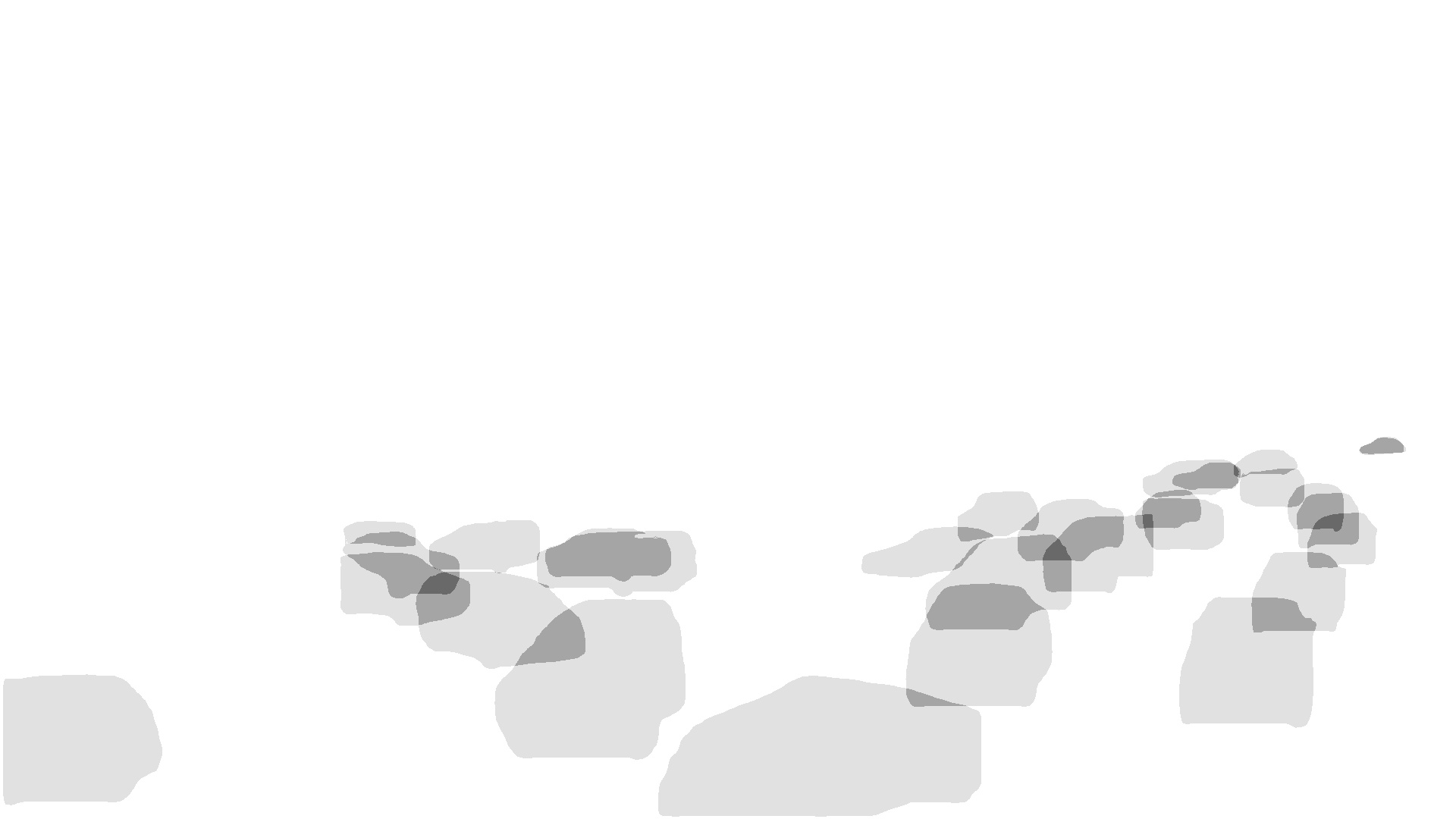}}
         \fbox{\includegraphics[width=0.232\textwidth,height=0.15\textwidth]{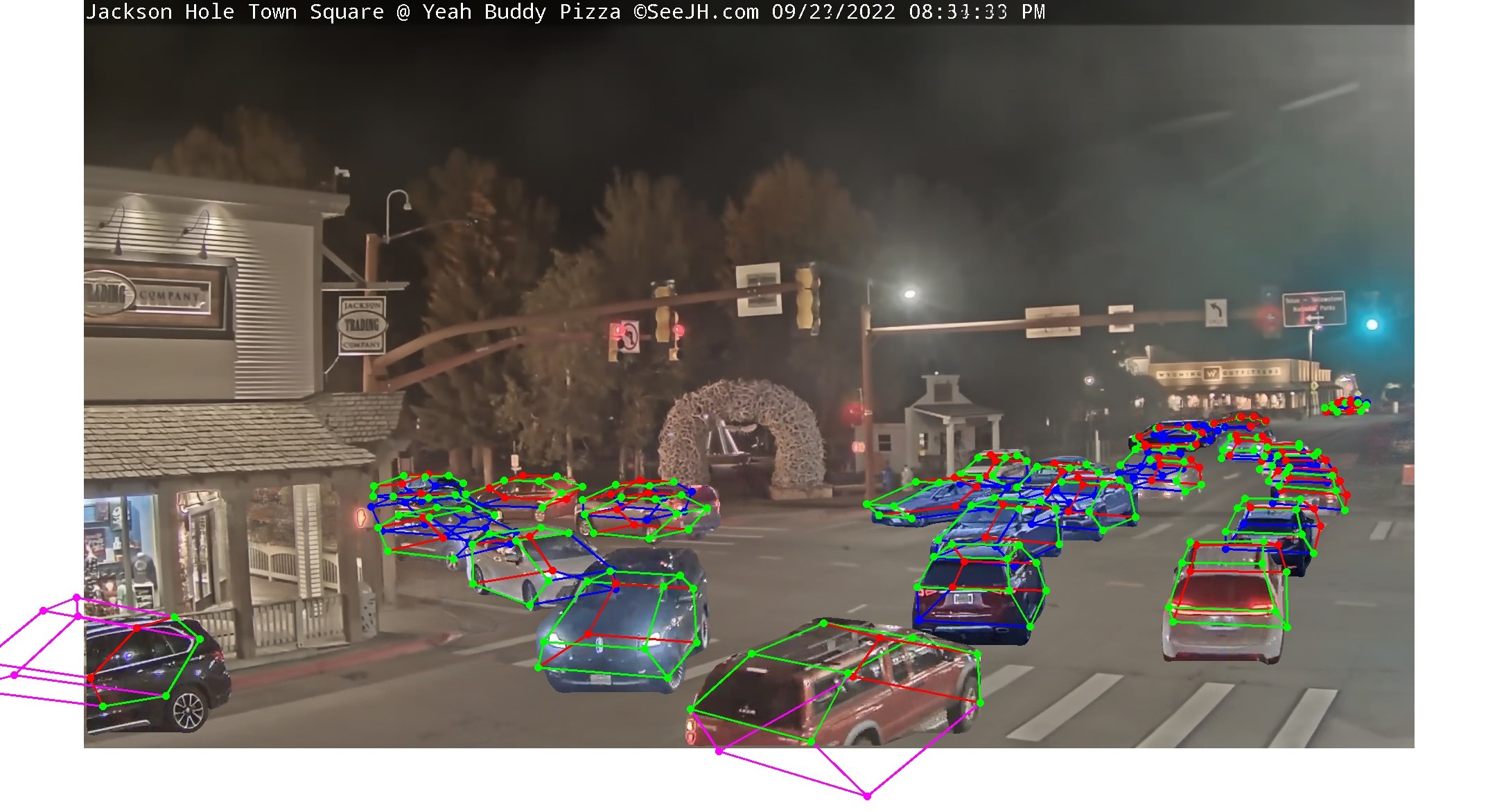}}
         \fbox{\includegraphics[width=0.232\textwidth,height=0.15\textwidth]{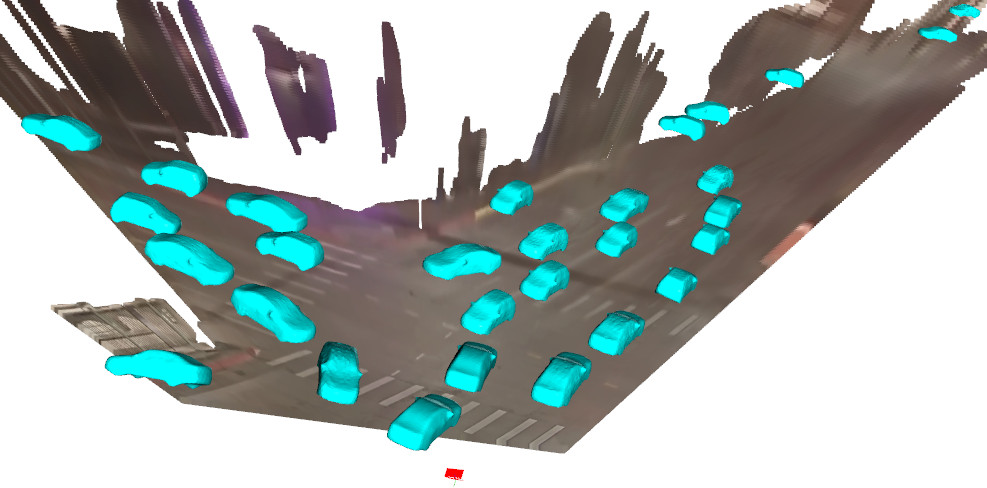}}

        \fbox{\includegraphics[width=0.232\textwidth,height=0.15\textwidth]{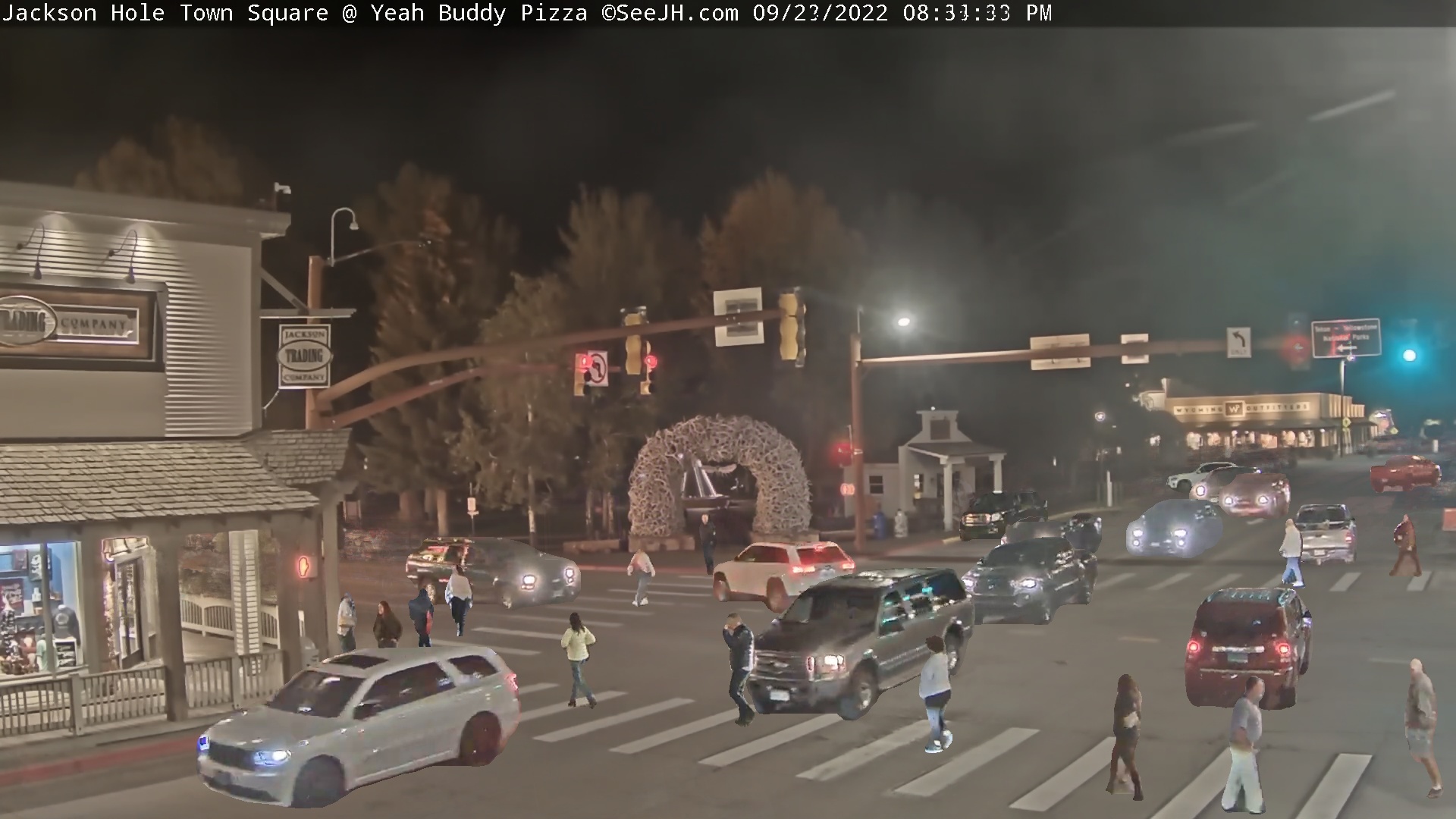}}
        \fbox{\includegraphics[width=0.232\textwidth,height=0.15\textwidth]{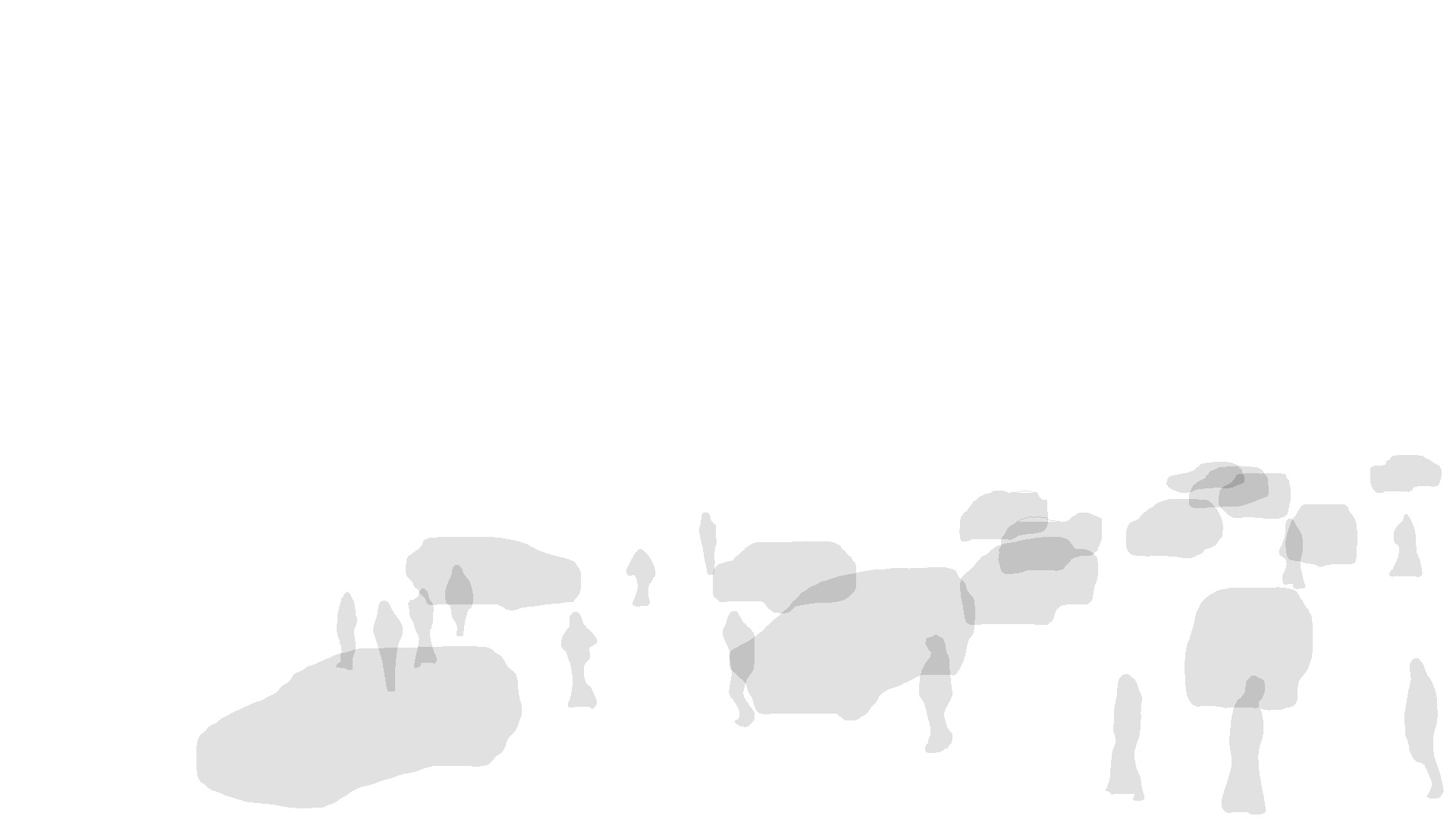}}
        \fbox{\includegraphics[width=0.232\textwidth,height=0.15\textwidth]{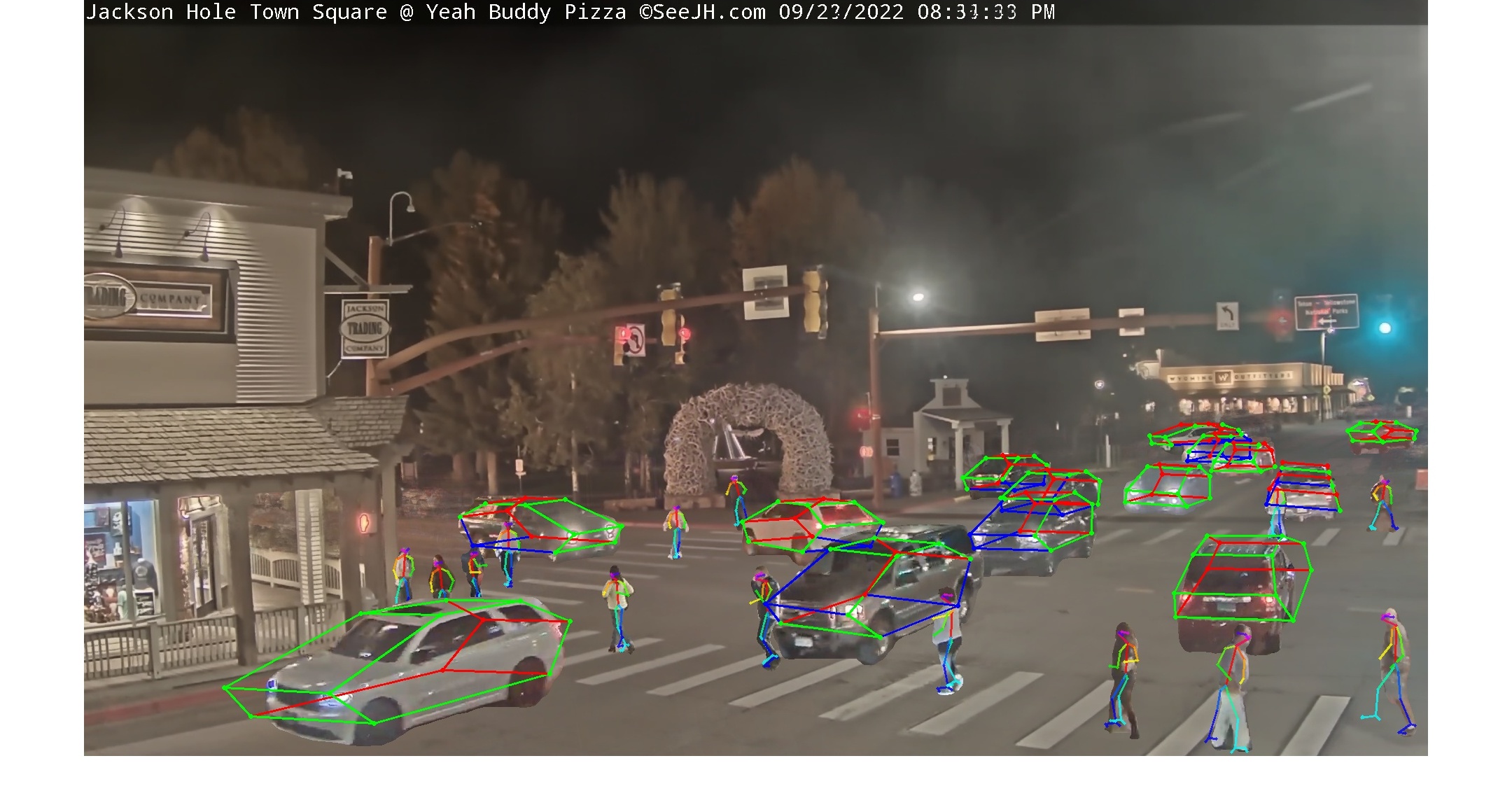}}
        \fbox{\includegraphics[width=0.232\textwidth,height=0.15\textwidth]{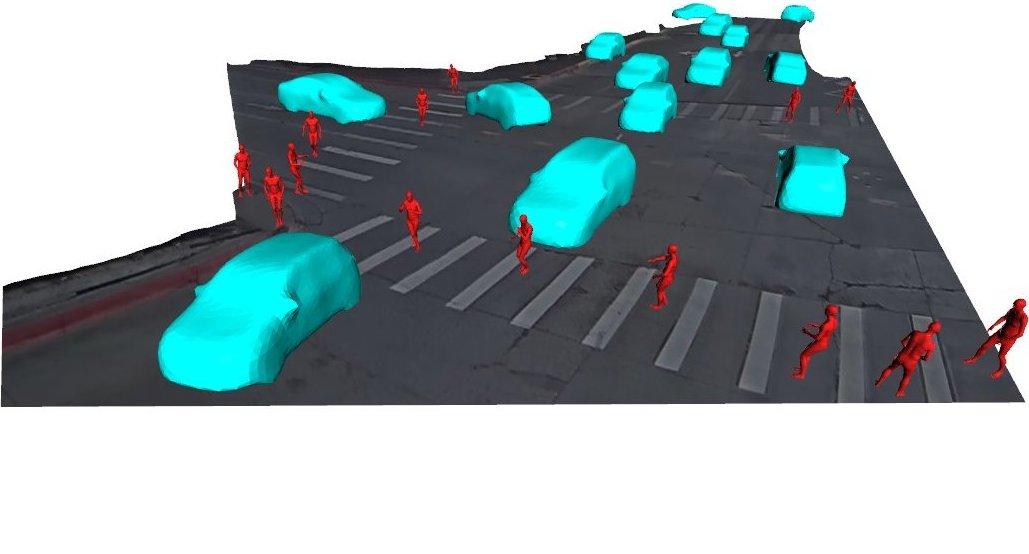}}

         \fbox{\includegraphics[width=0.232\textwidth,height=0.15\textwidth]{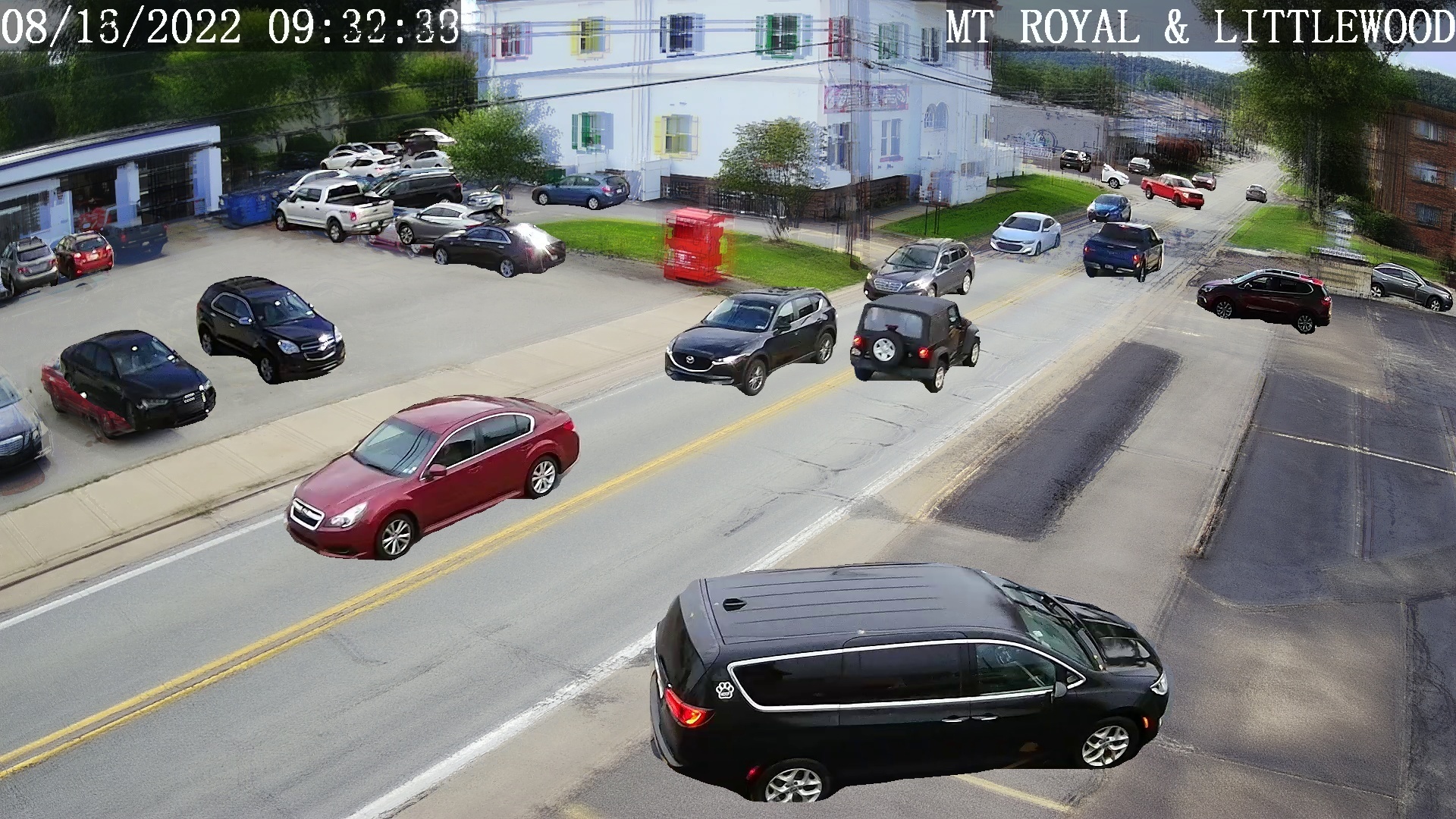}}
         \fbox{\includegraphics[width=0.232\textwidth,height=0.15\textwidth]{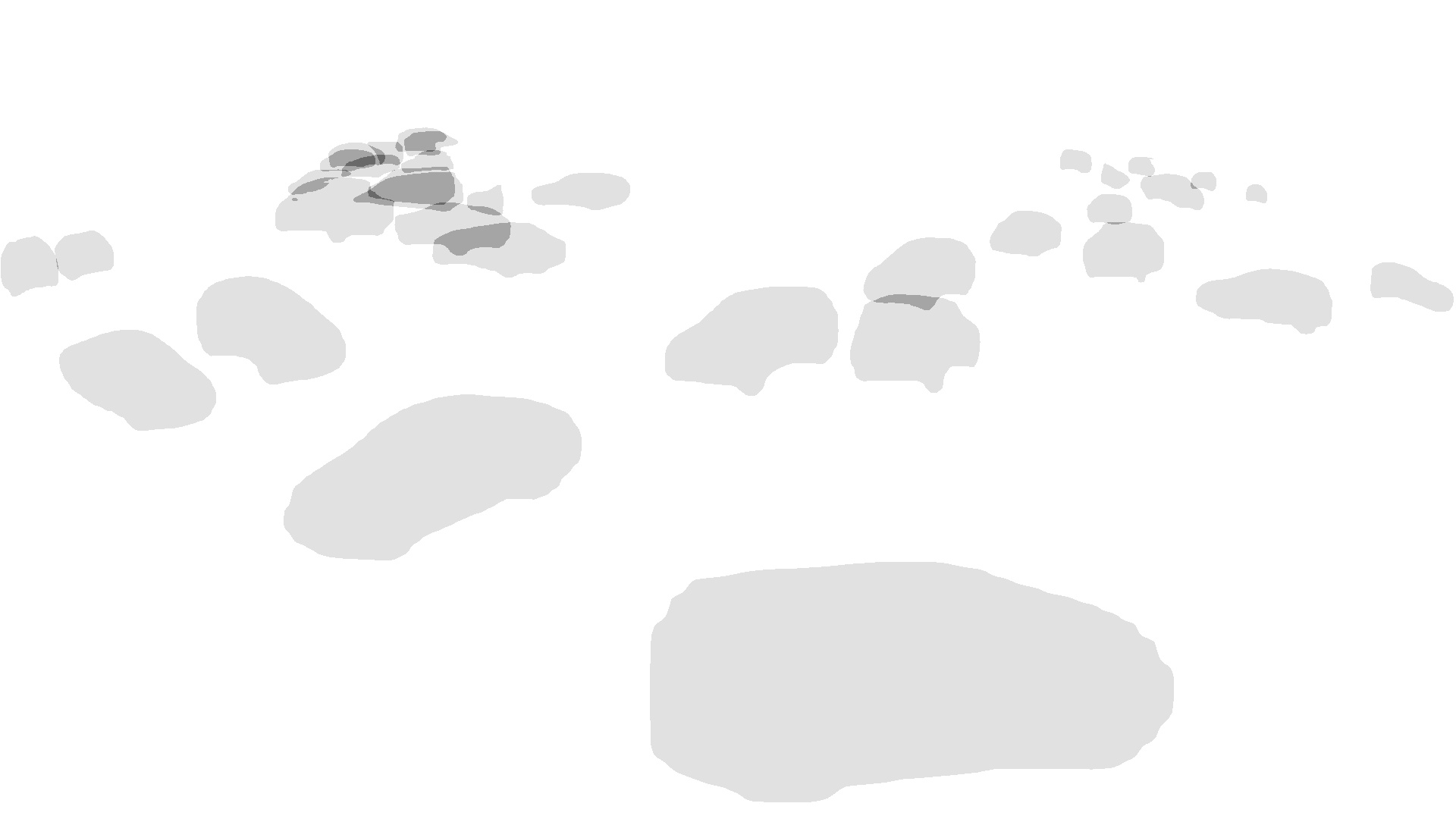}}
         \fbox{\includegraphics[width=0.232\textwidth,height=0.15\textwidth]{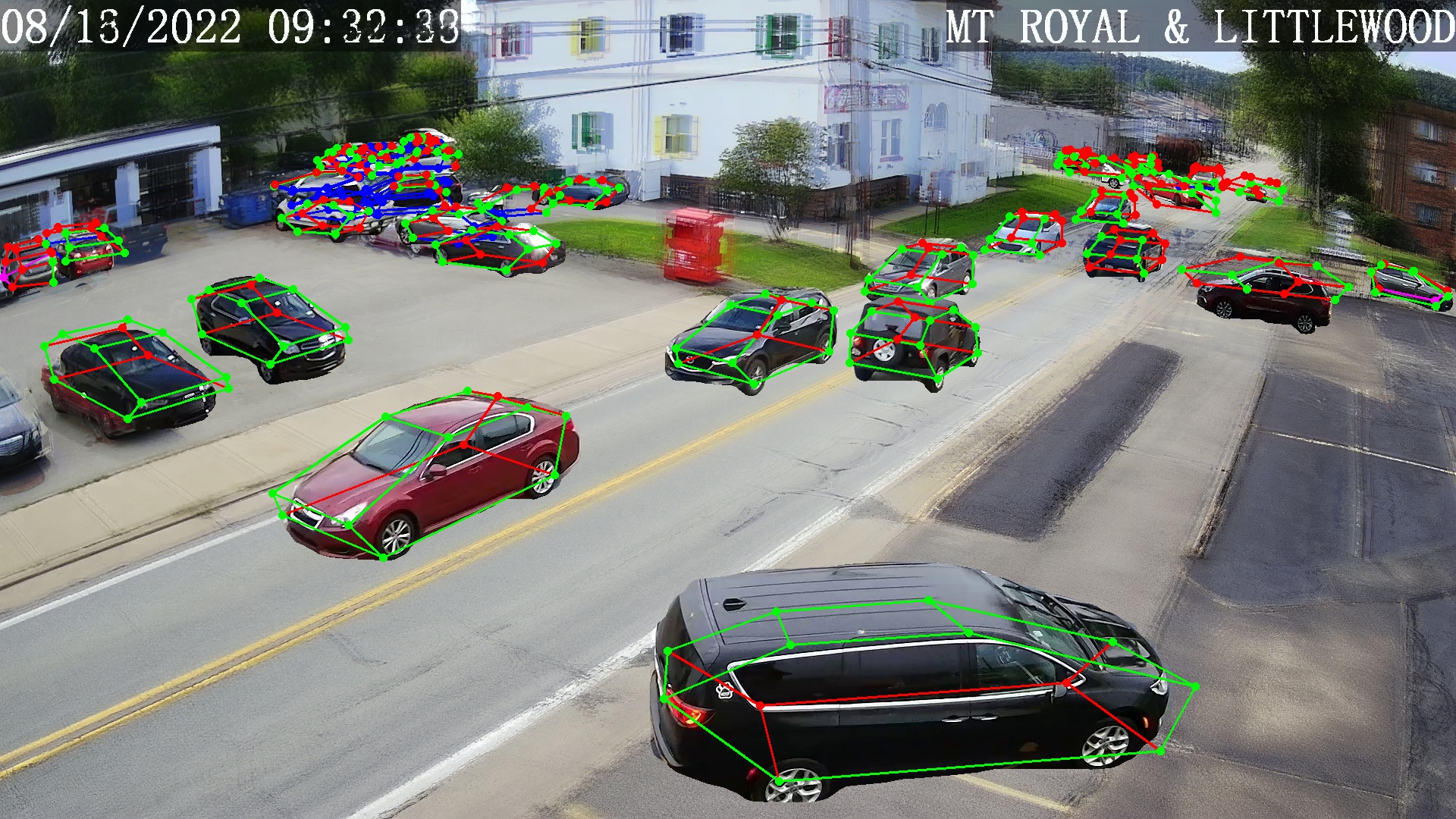}}
         \fbox{\includegraphics[width=0.232\textwidth,height=0.15\textwidth]{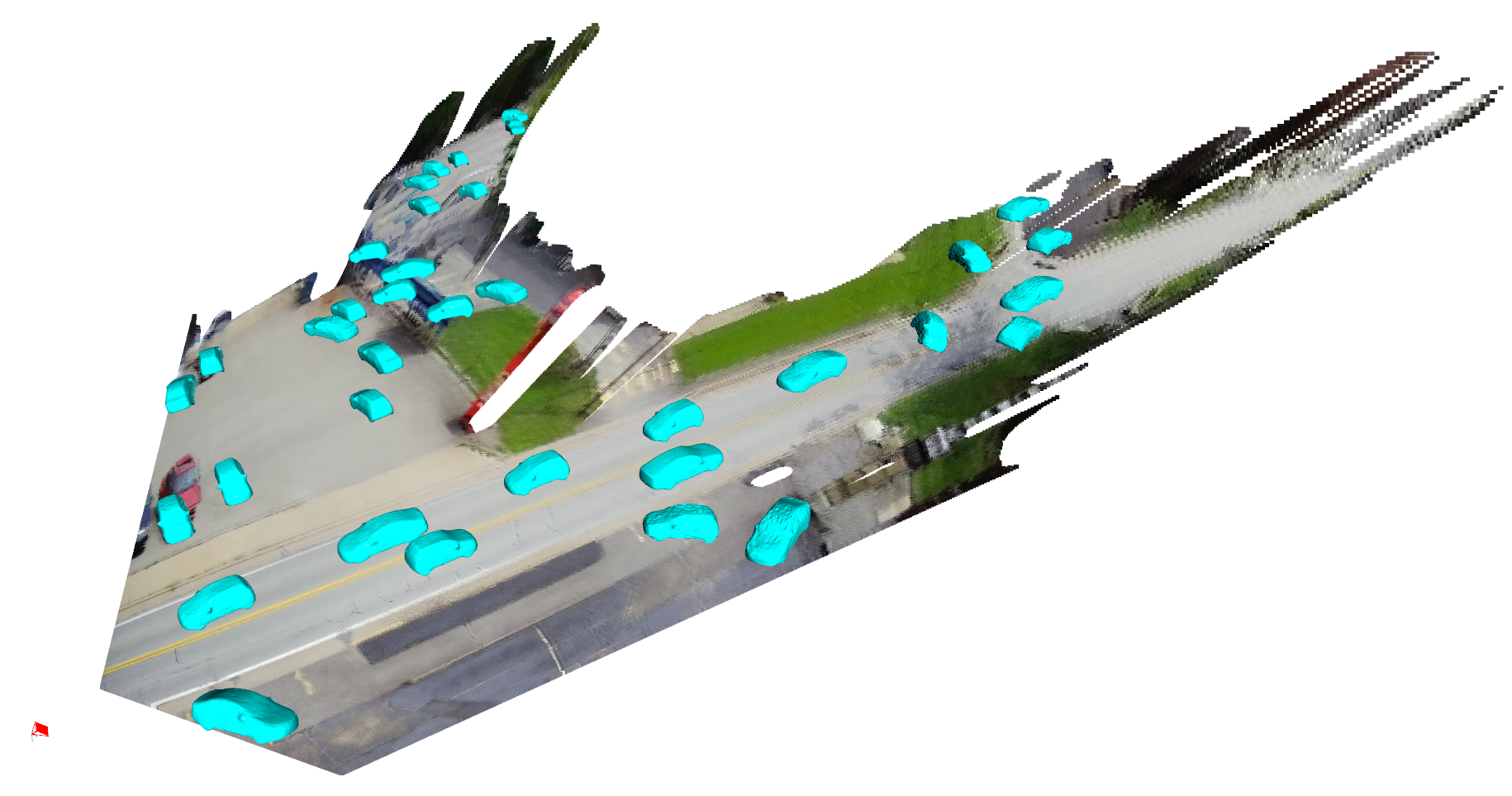}}
        
{\footnotesize         \textbf{Clip-Art Image \ \ \ \ \ \ \ \ \ \ \ \ \ \ \ \ \ \ \ \ \ \ \ \ \ \ \ \ \ \ \ \ \ \ \ \ \ \ \ Segmentation \ \ \ \ \ \ \ \ \ \ \ \ \ \ \ \ \ \ \ \ \  \ \ \ \ \ \ \ \ \ \ \ \ Keypoints \ \ \ \ \ \ \ \ \ \ \ \ \ \ \ \ \ \ \ \ \  \ \ \ \ \ \ \ \ \ \ \ \ \ \ \ \ \ \ \ 3D Reconstruction}}\\

\caption{{\bf Automatically generated 2D and 3D Clip-Art to supervise our network:} Unoccluded objects are first mined using time-lapse imagery of the WALT dataset \cite{Reddy_2022_CVPR}. Random non-intersecting unoccluded objects are composited back into the background image in their respective original positions to preserve correct appearances. The resulting Clip-Art images and their respective amodal segmentation masks, keypoint locations, and 3D meshes are shown. Our method generates realistic appearances from any camera, incorporating diverse viewing geometries, weather conditions, lighting, and occlusion configurations.}
    \label{fig:supp_clipart}
\end{figure*}

More examples from our 2D/3D Clip-Art pseudo-groundtruth supervision data, including the clip-art image with corresponding amodal segmentation, keypoints, and 3D object reconstruction, are shown in Fig.~\ref{fig:supp_clipart}.

\subsection{Additional Qualitative Results}
\begin{figure*}[h!]
    \centering


        \fbox{\includegraphics[width=0.18\textwidth,height=0.18\textwidth]{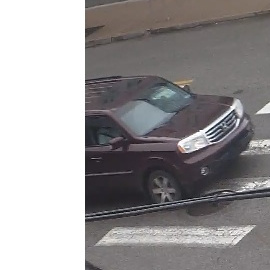}}
        \fbox{\includegraphics[width=0.18\textwidth,height=0.18\textwidth]{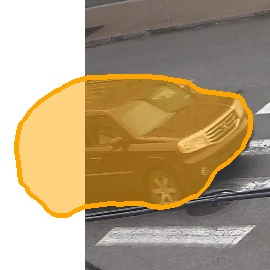}}
        \fbox{\includegraphics[width=0.18\textwidth,height=0.18\textwidth]{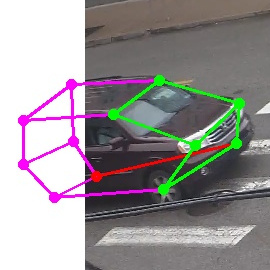}}
        \fbox{\includegraphics[width=0.18\textwidth,height=0.18\textwidth]{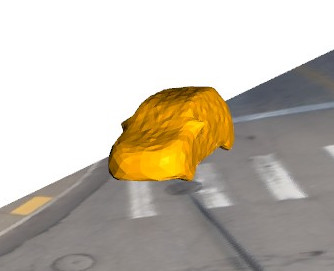}}
        \fbox{\includegraphics[width=0.18\textwidth,height=0.18\textwidth]{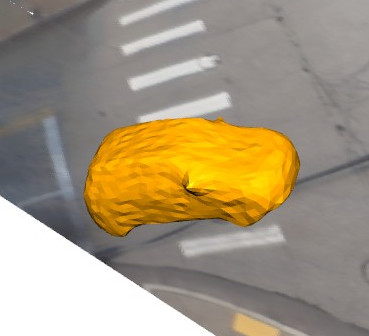}}

        \fbox{\includegraphics[width=0.18\textwidth,height=0.18\textwidth]{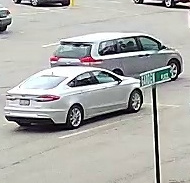}}
        \fbox{\includegraphics[width=0.18\textwidth,height=0.18\textwidth]{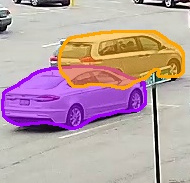}}
        \fbox{\includegraphics[width=0.18\textwidth,height=0.18\textwidth]{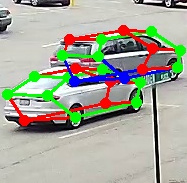}}
        \fbox{\includegraphics[width=0.18\textwidth,height=0.18\textwidth]{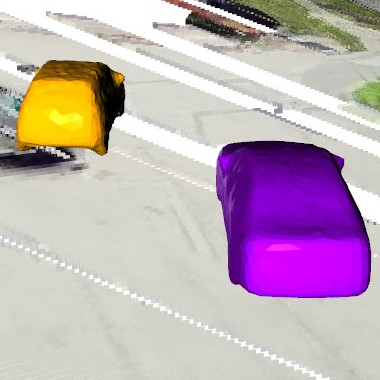}}
        \fbox{\includegraphics[width=0.18\textwidth,height=0.18\textwidth]{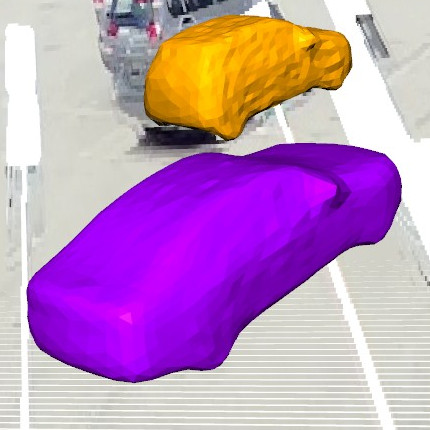}}

        \fbox{\includegraphics[width=0.18\textwidth,height=0.18\textwidth]{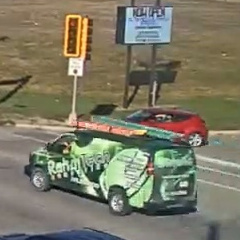}}
        \fbox{\includegraphics[width=0.18\textwidth,height=0.18\textwidth]{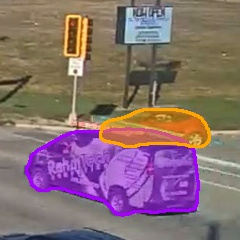}}
        \fbox{\includegraphics[width=0.18\textwidth,height=0.18\textwidth]{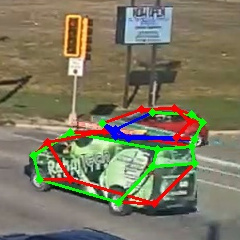}}
        \fbox{\includegraphics[width=0.18\textwidth,height=0.18\textwidth]{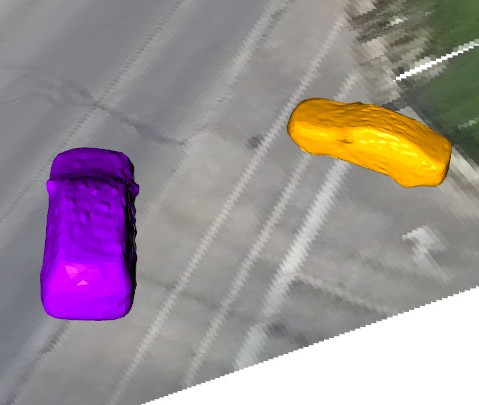}}
        \fbox{\includegraphics[width=0.18\textwidth,height=0.18\textwidth]{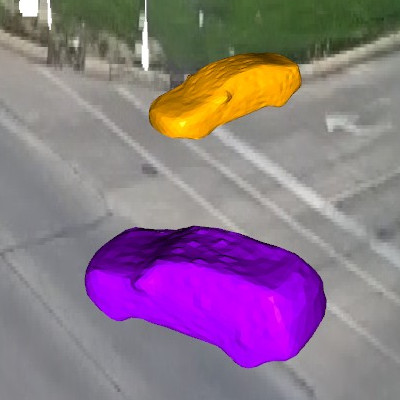}}

        \fbox{\includegraphics[width=0.18\textwidth,height=0.18\textwidth]{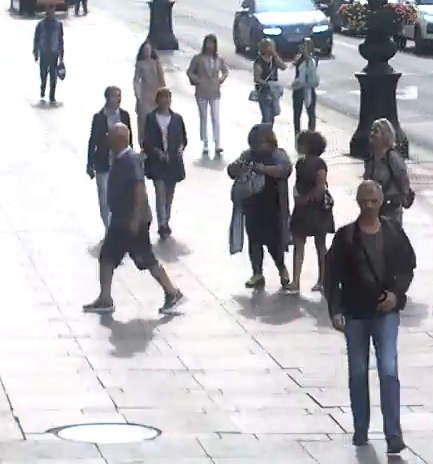}}
        \fbox{\includegraphics[width=0.18\textwidth,height=0.18\textwidth]{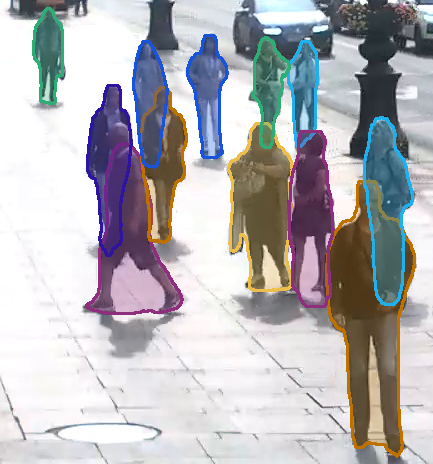}}
        \fbox{\includegraphics[width=0.18\textwidth,height=0.18\textwidth]{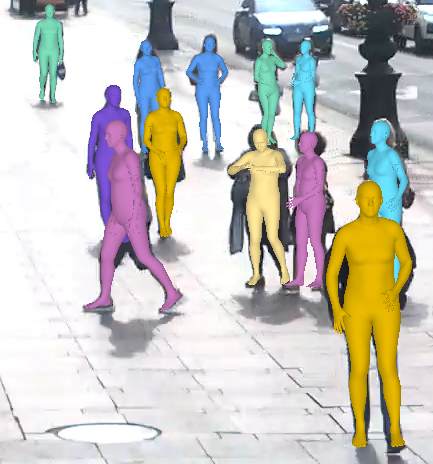}}
        \fbox{\includegraphics[width=0.18\textwidth,height=0.18\textwidth]{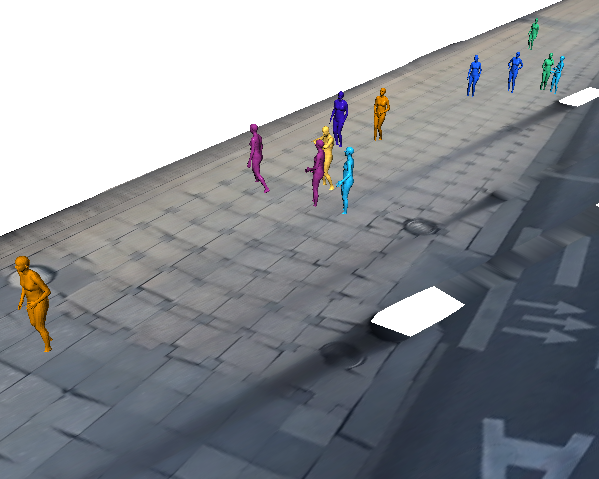}}
        \fbox{\includegraphics[width=0.18\textwidth,height=0.18\textwidth]{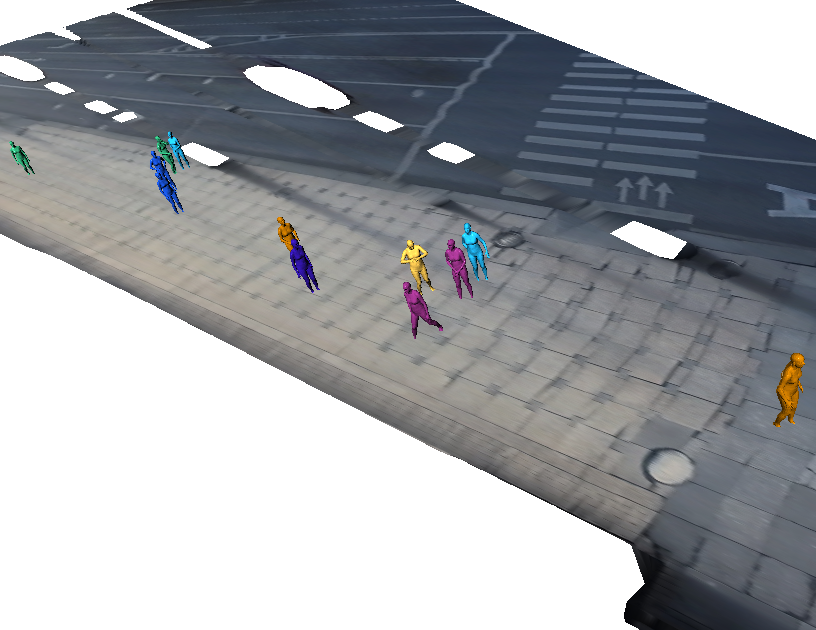}}

        \fbox{\includegraphics[width=0.18\textwidth,height=0.18\textwidth]{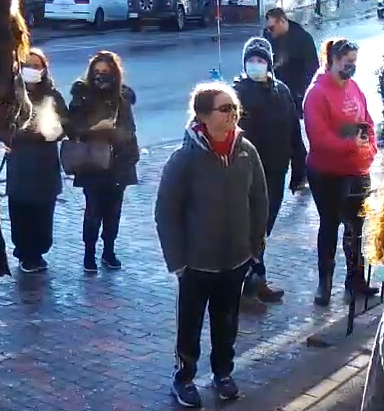}}
        \fbox{\includegraphics[width=0.18\textwidth,height=0.18\textwidth]{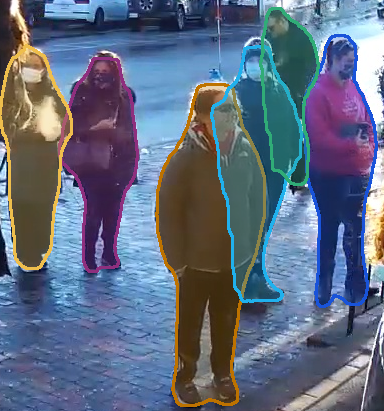}}
        \fbox{\includegraphics[width=0.18\textwidth,height=0.18\textwidth]{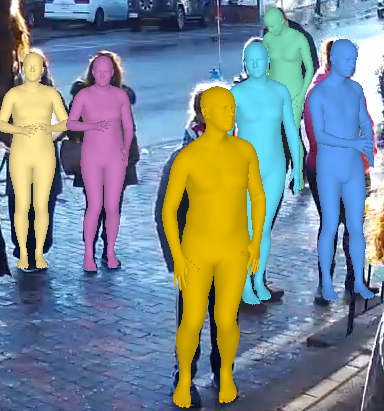}}
        \fbox{\includegraphics[width=0.18\textwidth,height=0.18\textwidth]{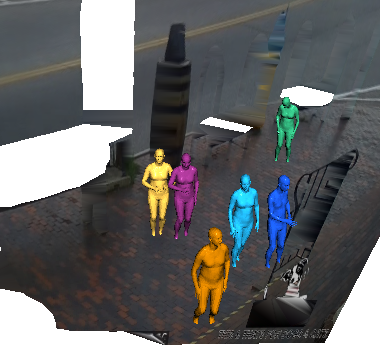}}
        \fbox{\includegraphics[width=0.18\textwidth,height=0.18\textwidth]{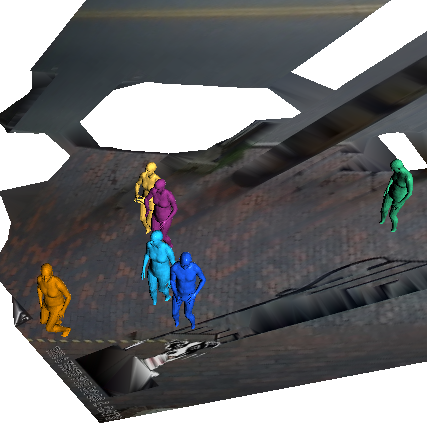}}
        
        \fbox{\includegraphics[width=0.18\textwidth,height=0.18\textwidth]{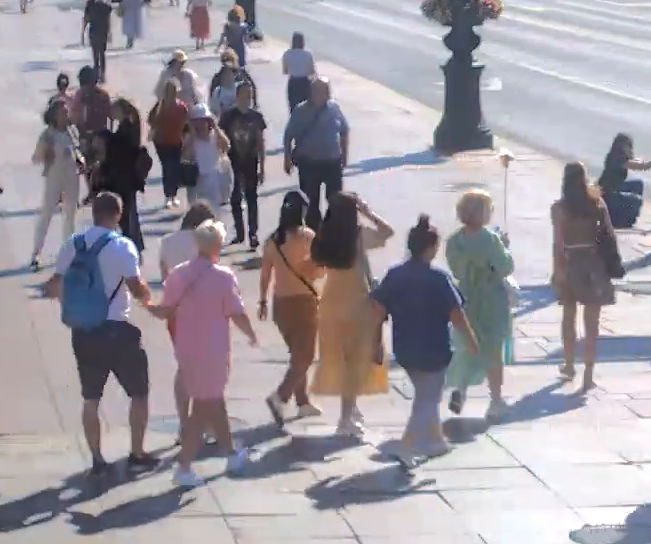}}
        \fbox{\includegraphics[width=0.18\textwidth,height=0.18\textwidth]{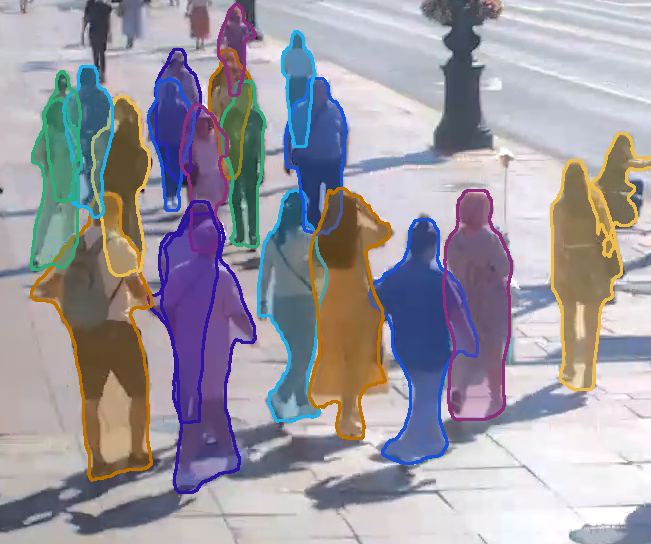}}
        \fbox{\includegraphics[width=0.18\textwidth,height=0.18\textwidth]{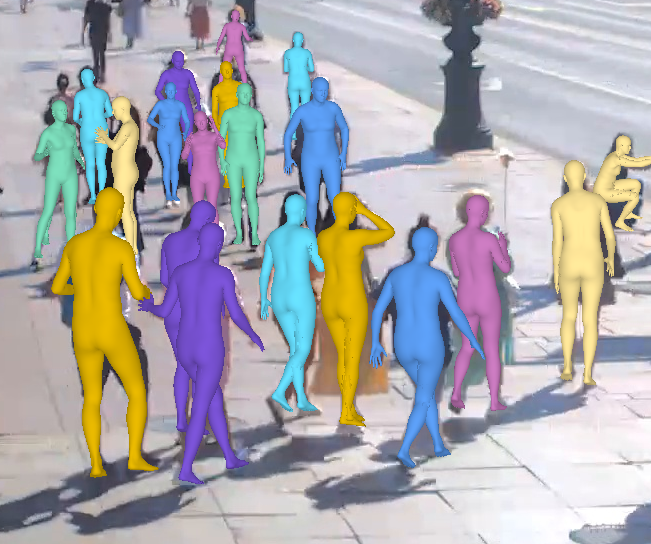}}
        \fbox{\includegraphics[width=0.18\textwidth,height=0.18\textwidth]{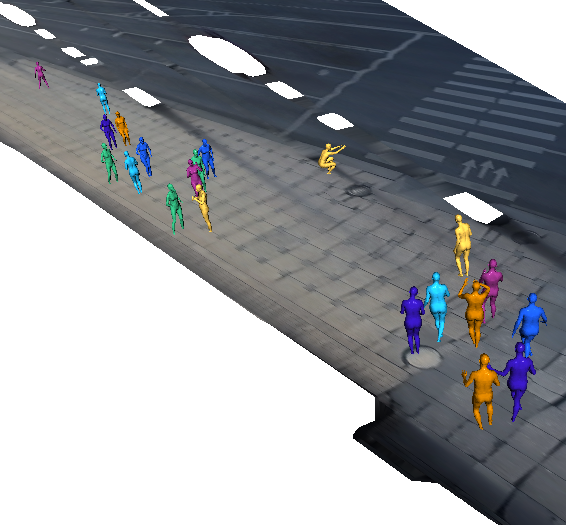}}
        \fbox{\includegraphics[width=0.18\textwidth,height=0.18\textwidth]{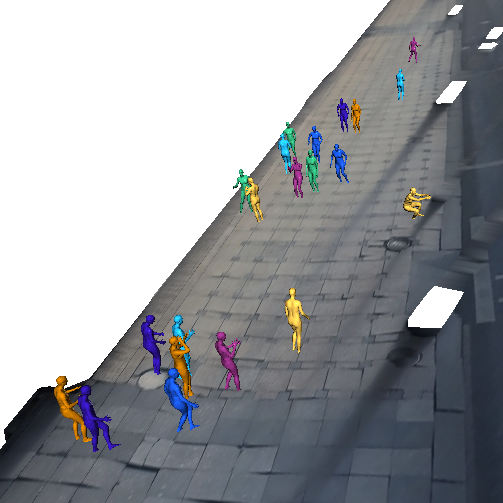}}

\textbf{\small \ \ \ \ Input Image \ \ \ \ \ \ \ \ \  \  \ \  Amodal Segmentation  \ \ \    Amodal Keypoints/Shapes  \  \ \ \ \ \ \ \ \ \ \ \ \ \    3D View-1  \ \ \ \ \ \ \ \ \ \ \ \  \ \ \ \ \ \ \ \ \  \ \ \ \ 3D View-2}

\caption{We show additional qualitative results on multiple sequences of the WALT~\cite{Reddy_2022_CVPR} dataset. Our method produces accurate amodal segmentation, keypoints, as well as 3D poses and shapes across diverse poses and occlusion configurations.}
    \label{fig:supp_final_results}
\end{figure*}

Additional results are shown in Fig.~\ref{fig:supp_final_results}, with various occlusion configurations, including self-occlusion, truncation, and occlusion-by-others. Notably, training with our clip-art data yields a substantial improvement over baseline methods, particularly in scenarios with heavy occlusion.

\end{document}